\documentclass{article} 
\usepackage{iclr2025_conference,times}

\usepackage{amsmath,amsfonts,bm}

\def\eqref#1{equation~\ref{#1}}

\def\1{\bm{1}}

\def\mH{{\bm{H}}}

\DeclareMathAlphabet{\mathsfit}{\encodingdefault}{\sfdefault}{m}{sl}
\SetMathAlphabet{\mathsfit}{bold}{\encodingdefault}{\sfdefault}{bx}{n}

\def\gG{{\mathcal{G}}}

\usepackage{graphicx}
\usepackage{xcolor}
\usepackage{multirow}
\usepackage{hyperref}
\usepackage{url}
\usepackage{wrapfig}

\usepackage{amsthm}
\newtheorem{theorem}{Theorem}[section]

\usepackage{etoc}
\etocdepthtag.toc{mtchapter}
\etocsettagdepth{mtchapter}{subsection}
\etocsettagdepth{mtappendix}{none}

\newenvironment{myquotation}{\setlength{\leftmargini}{0em}\quotation}{\endquotation}

\title{BrainOOD: Out-of-distribution Generalizable Brain Network Analysis}

\author{%
  \textbf{Jiaxing Xu$^1$\footnotemark[1] , Yongqiang Chen$^2$\thanks{Equal contribution} ,  Xia Dong$^1$, Mengcheng Lan$^3$, Tiancheng Huang$^1$, } \\ \textbf{Qingtian Bian$^1$, James Cheng$^2$, Yiping Ke$^1$}\\
  $^1$College of Computing and Data Science, Nanyang Technological University\\
  $^2$Department of Computer Science and Engineering, The Chinese University of Hong Kong\\
  $^3$S-Lab, Nanyang Technological University\\
\texttt{\{jiaxing003, LANM0002, BIAN0027\}@e.ntu.edu.sg}; \\
\texttt{\{yqchen, jcheng\}@cse.cuhk.edu.hk};\\ 
\texttt{\{xia.dong, tiancheng.huang, ypke\}@ntu.edu.sg} \\
}

\iclrfinalcopy 
\begin{document}

\maketitle

\begin{abstract}
In neuroscience, identifying distinct patterns linked to neurological disorders, such as Alzheimer’s and Autism, is critical for early diagnosis and effective intervention. Graph Neural Networks (GNNs) have shown promising in analyzing brain networks, but there are two major challenges in using GNNs: (1) distribution shifts in multi-site brain network data, leading to poor Out-of-Distribution (OOD) generalization, and (2) limited interpretability in identifying key brain regions critical to neurological disorders. Existing graph OOD methods, while effective in other domains, struggle with the unique characteristics of brain networks. To bridge these gaps, we introduce \textit{BrainOOD},  a novel framework tailored for brain networks that enhances GNNs’ OOD generalization and interpretability. BrainOOD framework consists of a feature selector and a structure extractor, which incorporates various auxiliary losses including an improved Graph Information Bottleneck (GIB) objective to recover causal subgraphs. By aligning structure selection across brain networks and filtering noisy features, BrainOOD offers reliable interpretations of critical brain regions. Our approach outperforms 16 existing methods and improves generalization to OOD subjects by up to 8.5\%. Case studies highlight the scientific validity of the patterns extracted, which aligns with the findings in known neuroscience literature. We also propose the first OOD brain network benchmark, which provides a foundation for future research in this field. Our code is available at \url{https://github.com/AngusMonroe/BrainOOD}.
\end{abstract}

\section{Introduction}
\label{sec:intro}
In neuroscience, a major goal is to identify distinct patterns linked to neurological disorders, such as Alzheimer’s and Autism, by examining brain data of both healthy individuals and patients with these disorders \citep{poldrack2009decoding}. 
Among the neuroimaging techniques, resting-state functional magnetic resonance imaging (fMRI) is widely used to capture the functional connectivity between different brain regions \citep{worsley2002general}. 
fMRI can be modeled as brain networks, where each node represents a brain region, referred to as a region of interest (ROI), and each edge denotes the pairwise correlation between the blood-oxygen-level-dependent (BOLD) signals of two ROIs \citep{smith2011network}. These connections provide insights into how different brain regions co-activate or show correlated activities, offering a framework to study neurological systems through graph-based methods \citep{kawahara2017brainnetcnn,lanciano2020explainable,wang2023effective,xu2024multi}. 

The most prevalent brain network analysis model is based on Graph Neural Networks (GNNs), which have recently shown promising results~\citep{li2019graph,li2021braingnn,xu2024contrastive}. However, the application of GNN-based methods in brain network analysis poses two significant challenges. First, brain network data are often collected from different sites, leading to \textit{distribution shifts}, which severely degrade the performance of GNNs when generalizing to Out-of-Distribution (OOD) data during testing~\citep{ciga,xu2024contrasformer}. Second, brain network analysis aims to uncover patterns that can facilitate early diagnosis and interventions for neurological disorders. This requires GNN models to possess strong \textit{interpretability}, allowing them to identify key brain regions relevant to the concerned conditions.

\begin{wrapfigure}{r}{0.3\textwidth}
\vspace{-10pt}
\begin{center}
\includegraphics[width=\linewidth]{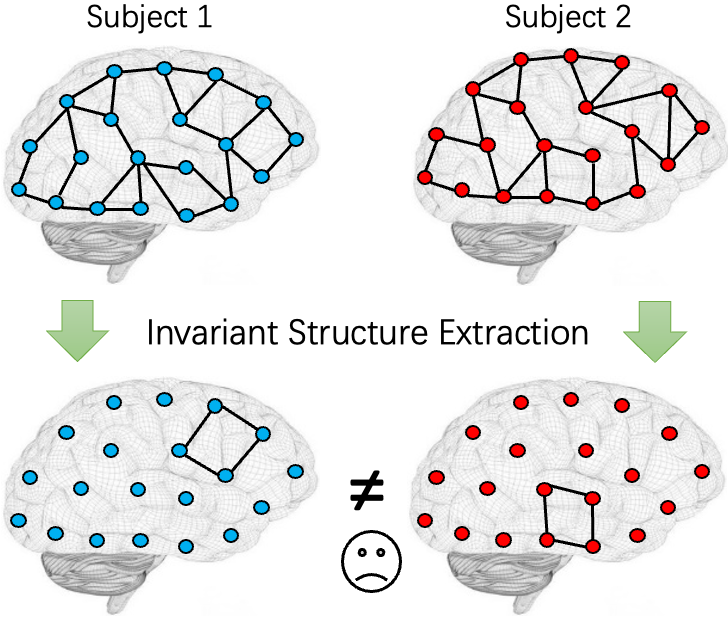}
\end{center}
\vspace{-6pt}
\caption{Same substructure in different brain regions may reflect distinct functional implications.}
\vspace{-12pt}
\label{fig:example}
\end{wrapfigure}

Several interpretable GNN methods~\citep{wu2022dir,miao2022interpretable,cheninterpretable} have been proposed to address the OOD generalization problem. These methods assume that a causal subgraph contains the essential information for predictions, thus improving the robustness to distribution shifts. While this is effective in domains like molecular and social networks, such an approach struggles with brain network analysis. Unlike other graph-structured data, brain networks exhibit noise in both their structures and features. Existing methods primarily focus on extracting causal substructures, often overlooking the selection of critical node features, which limits their applicability to brain networks. Additionally, invariant subgraphs identified by these methods may not effectively interpret brain networks. As shown in Figure \ref{fig:example}, invariant substructures involving different ROIs can reflect distinct functional implications, highlighting the need for a specialized approach. This leads to a key research question:

\begin{myquotation}\centering
    \textit{How can one build an interpretable and OOD generalizable brain GNN?}
\end{myquotation}

To tackle the aforementioned challenges, we propose the first benchmark dataset for evaluating OOD  generalization performance in brain network analysis. Specifically, we go beyond the conventional usage of brain network datasets by creating a specific OOD benchmark scenario that simulates real-world conditions where models encounter data from unseen sites during testing. Building on this benchmark, we develop \textit{BrainOOD}, which is a novel framework that enhances GNNs' representation power and enables the recovery of causal subgraphs using an improved Graph Information Bottleneck (GIB) objective \citep{wu2020graph}. 
BrainOOD includes a feature selector and a structure extractor. The feature selector introduces a learnable masking process to selectively filter out noisy node features. A high-pass GNN with a reconstruction objective is incorporated to recover informative node features and learns high-quality representations to reveal causally interpretable brain regions. Additionally, we adopt a discrete sampling strategy for structure extraction. This ensures the identification of critical connections and enforces alignment across samples for consistent structure selection.
Our contributions are summarized as follows:

\begin{itemize}
    \item We introduce the first benchmark for evaluating OOD performance in brain network analysis. Our proposed benchmark is the first to systematically evaluate OOD generalization on brain network datasets with a focus on addressing site-specific variability, which is a critical challenge in clinical applications.
    \item We propose BrainOOD, a novel architecture that enhances OOD generalization in brain networks by selectively extracting node features and graph structures, while exploiting the inherent node alignment in brain networks.
    \item We evaluate BrainOOD against 16 existing methods and demonstrate its superior performance, improving generalization to OOD subjects by up to 8.5\%.
    \item We present a case study to showcase the highly interpretable and scientifically meaningful patterns identified by BrainOOD, which align with the findings in neuroscience literature.
\end{itemize}

\section{Preliminaries}
\label{sec:preliminaries}

\subsection{Brain Networks Classification}

We use the brain networks released by \citet{xu2023data}.
All preprocessed fMRI are parcellated by Schaefer atlas with 100 ROIs \citep{schaefer2018local}.
For each subject, a brain network was constructed in the form of a connectivity matrix, $\boldsymbol{S}$, where the nodes represent ROIs, and the edges encode Pearson's correlation between the region-averaged BOLD signals of each pair of ROIs. Essentially, $\boldsymbol{S}$ captures the functional relationships between different brain regions. To represent the brain network as a graph $G = (\boldsymbol{X}, \boldsymbol{A})$, we define the feature matrix $\boldsymbol{X} = \boldsymbol{S}$, and the adjacency matrix $\boldsymbol{A}$ as a sparsified version of $\boldsymbol{S}$, retaining the top 20\% of connections with the highest correlations. Notably, by using a consistent parcellation method, all brain networks share the same number of nodes $n = 100$, corresponding to the fixed set of ROIs.

Brain network classification aims to predict a subject's condition (e.g., autism diagnosis) based on his/her brain network. Given a dataset $\mathcal{D} = (\mathcal{G}, \mathcal{Y}) = \{(G, y_G)\}$, where $G \in \mathcal{G}$ represents a brain network and $y_G$ is its corresponding class label, the task is to learn a predictive function $f$: $\mathcal{G} \rightarrow \mathcal{Y}$ that maps brain networks to their respective labels.
In this work, our objective of brain network classification is not only to accurately classify the networks in the training dataset but also to ensure that the learned function $f$ generalizes well to unseen or OOD brain networks, which may come from different sites with different feature distributions.
In addition to the OOD generalizable predictions, we also aim to provide meaningful interpretations for the predictions by identifying a subgraph $G_C$ of the input brain network, offering insights into the functionalities of different ROIs.
We summarize the notations used throughout the paper in Appendix \ref{app:notation}.

\subsection{Graph Neural Networks (GNNs)} 

GNNs have emerged as powerful tools for brain network analysis due to their ability to incorporate both node attributes and topological structures. Consider an input graph $G = (\boldsymbol{X}, \boldsymbol{A})$, where $\boldsymbol{A}$ is the adjacency matrix, which encodes connectivity information, and $\boldsymbol{X}$ is the feature matrix containing attribute information for each node. The node set of $G$ is denoted as $\mathcal{V}_G$ and $|\mathcal{V}_G| = n$.
The $l$-th layer of a GNN in the message-passing scheme \citep{xu2018powerful} can be written as: 
\begin{equation}
\label{eq:mpnn}
\small
\boldsymbol{H}_{v}^{(l)}\!=\!\operatorname{AGG}^{(l-1)}\!\left(\boldsymbol{H}_{v}^{(l-1)}, \operatorname{MSG}^{(l-1)}\!\left(\left\{\!\boldsymbol{H}_{u}^{(l-1)}\!\right\}_{\forall u \in \mathcal{N}(v)}\!\right)\!\right),
\end{equation}
where $\boldsymbol{H}^{(l)}_v \in \mathbb{R}^{d}$ denotes the node representation at the $l$-th layer, where each node is represented by a $d$-dimensional vector. $\operatorname{AGG}(\cdot)$ and $\operatorname{MSG}(\cdot)$ are arbitrary differentiable aggregate and message functions (e.g., a multilayer perceptron (MLP) can be used as $\operatorname{AGG}(\cdot)$ and a summation function as $\operatorname{MSG}(\cdot)$). $\mathcal{N}(v)$ represents the neighbor node set of node $v \in \mathcal{V}_G$, and $\boldsymbol{H}^{(0)}_v = \boldsymbol{X}_v$ representing the raw features of node $v$. 


In contrast to conventional message-passing GNNs, where information is aggregated from a node’s neighbors, a high-pass graph neural network (HPGNN) emphasizes the differences between a node's features and the aggregated features of its neighbors. This approach is especially useful for capturing local variations in brain networks. The update rule for an HPGNN layer is:
\begin{equation}
\label{eq:HPGNN}
\small
\boldsymbol{H}_{v}^{(l)} = \boldsymbol{H}_{v}^{(l-1)} - \operatorname{AGG}^{(l-1)}\left(\operatorname{MSG}^{(l-1)}\left(\left\{\boldsymbol{H}_{u}^{(l-1)}\right\}_{\forall u \in \mathcal{N}(v)}\right)\right).
\end{equation} 
This operation enables the model to focus on deviations from local patterns, which may be critical in detecting abnormal or OOD graph substructures.

\section{Out-of-distribution Benchmark in Brain Network Analysis}
\label{sec:benchmark}

\subsection{Distribution Shifts in Brain Network Analysis}

One of the primary goals in analyzing neurological disorders is to uncover disease-specific patterns that remain consistent across diverse populations. However, brain network datasets often exhibit distribution shifts \citep{xu2024contrasformer}, where features common to specific sub-populations are mistakenly identified as disease-related, despite being unrelated to the disorder. This can result in models learning spurious connections that do not generalize across the broader population.
For instance, large-scale brain network datasets like the Autism Brain Imaging Data Exchange (ABIDE) \citep{craddock2013neuro} and the Alzheimer’s Disease Neuroimaging Initiative (ADNI) \citep{dadi2019benchmarking} are collected from multiple sites, such as various clinics or universities. Subjects from these different sites may introduce site-specific variability, such as differences in MRI scanner properties, or subject inclusion/exclusion criteria \citep{chan2022semi}. Such factors contribute to site-specific biases, where models inadvertently focus on site-related patterns rather than capturing population-invariant information about the disorders.
The presence of this type of noise poses a significant challenge for model generalization, particularly in real-world medical applications, where deployment environments are rarely identical to training settings. Understanding and addressing these distribution shifts is crucial for improving the robustness and generalizability of brain network analysis models.
 
\subsection{Dataset under OOD Setting}

In medical applications, models are often trained on data collected from a limited number of sites but are expected to perform well across different, unseen sites during deployment. This scenario introduces OOD challenges, as variations between training and deployment sites can significantly degrade model performance.
To investigate this OOD shift, we use two widely-studied, multi-site brain network datasets:
ABIDE \citep{craddock2013neuro}, focused on Autism Spectrum Disorder (ASD), and
ADNI \citep{dadi2019benchmarking}, centered around Alzheimer’s Disease (AD).
The statistics of these datasets are summarized in Table \ref{tab:dataset_statistic}, and further detailed descriptions are provided in Appendix \ref{app:dataset}. Both datasets were collected from multiple sites, with inherent inter-site variability in acquisition and processing methods. This variability provides an ideal testbed for evaluating model performance under OOD conditions.
To simulate an OOD setting, we adopt a site-holdout strategy: each dataset is split into training, validation, and test sets in an 8:1:1 ratio. Importantly, the validation/test set is composed entirely of subjects from one specific site that were not present in the training set, making them OOD samples relative to the training data. This setup simulates the real-world scenario where a model trained on data from one set of sites is deployed in new, unseen environments. A detailed description of data split is included in Appendix \ref{app:split}.
For model evaluation, we use a consistent random seed across all experiments and perform 10-fold cross-validation. The average accuracy across folds is reported to ensure robustness in the results, allowing us to fairly compare models' generalization performance under OOD conditions.

\begin{table}[h]
\vspace{-0.1in}
\centering
\small
\caption{Statistics of Brain Network Datasets.}
\setlength\tabcolsep{4.5pt}
\begin{tabular}{cccccc}
\hline
Dataset & Condition              & Subject\#    & Site\#    & Class\#  & Class Name \\ \hline
ABIDE   & Autism Spectrum   Disorder           & 1025   & 17   & 2  & \{TC, ASD\}     \\
ADNI    & Alzheimer’s   Disease                & 1326   & 59   & 6    & \{CN, SMC, EMCI, MCI, LMCI, AD\}   \\ \hline
\end{tabular}
\label{tab:dataset_statistic}
\vspace{-0.1in}
\end{table}

\section{BrainOOD}
\label{sec:method}
Brain networks differ from regular graph data in that the co-activity representations in brain networks can contain a lot of noise. Meanwhile, the interpretable biomarkers in brain network analysis are usually similar for the same target disorder. Therefore, it brings additional challenges in data modeling and objective design.
In this section, we first demonstrate the failure of the existing GIB-based method and then propose several strategies to tackle the challenges.

\subsection{Interpretable and Generalizable Brain Network Analysis}
\label{subsec:gib}

In this work, our objective is to propose a robust GNN framework that can accurately predict the targets under distribution shifts.
Meanwhile, we also aim to identify a subregion in brain networks to explain the target analysis results such as Autism Spectrum Disorder and Alzheimer’s Disease, therefore, providing insights for future scientific discoveries. 

Specifically, we adopt the Graph Information Bottleneck (GIB) framework~\citep{wu2020graph,miao2022interpretable,cheninterpretable}, which can be formulated as follows:
\begin{equation}\label{eq:gib}
\small
    \text{$\max$}_{G_C}I(G_C; y_G)-\beta I(G_C;G),\ G_C\sim g_\phi(G),
\end{equation}
where $G_C$ encapsulates the causal information in $G$ that determines the target label $y_G$, $\beta\in[0,1]$ is a trade-off hyperparameter, $g_\phi:\gG\rightarrow\mathbb{G}(\gG)$ is the subgraph extractor parameterized by $\phi$, $\mathbb{G}(\gG)$ refers to the space of subgraphs for $G\in\gG$, and $I(\cdot;\cdot)$ is the mutual information.
\citet{ciga,miao2022interpretable,cheninterpretable} show that GIB can effectively solve for the desired causal subgraph $G_C^*$ in accordance with  Eq.~(\ref{eq:gib}) under distribution shifts.

However, when applying GIB to brain networks, several new challenges arise: (a) \textbf{low informative features}, as the node features and connections refer to the co-occurrence of brain activities in different ROIs; and (b) \textbf{unified interpretation}, as the interpretable ROIs for all subjects under the same condition should be similar. 

Consequently, the expressiveness and the representational power of GNNs can be further limited when used to seek interpretable ROIs under the aforementioned constraints. 
The limited representational power of GNNs will further lead to suboptimal generalization and interpretations. More formally, we have the following theorem:
\begin{theorem}\label{thm:gib_loss}
    For a subgraph extractor $g_\phi$ that encodes the input graph $G$ into representation $\mH$ to extract the desired subgraph $G_C^*$, if $g_\phi$ is limited in representation power, i.e., $I(G;\mH)< H(G_C^*)$, where $H(\cdot)$ is the entropy of the underlying causal subgraph $G_C^*$, then solving for GIB objective (Eq.~(\ref{eq:gib})) can not elicit $G_C^*$.
\end{theorem}
The proof is given in Appendix~\ref{proof:gib_loss}.
Theorem~\ref{thm:gib_loss} implies that it is essential to enhance the representation power of $g_\phi$ to effectively uncover the desired causal subgraph $G^*_C$. Consequently, we propose a new framework aimed at maximizing $I(G; H)$, while simultaneously incorporating an interpretation consistency regularization that ensures the structure of $G_C$ remains consistent across different samples.

\begin{figure*}[h]
\begin{center}
\includegraphics[width=\linewidth]{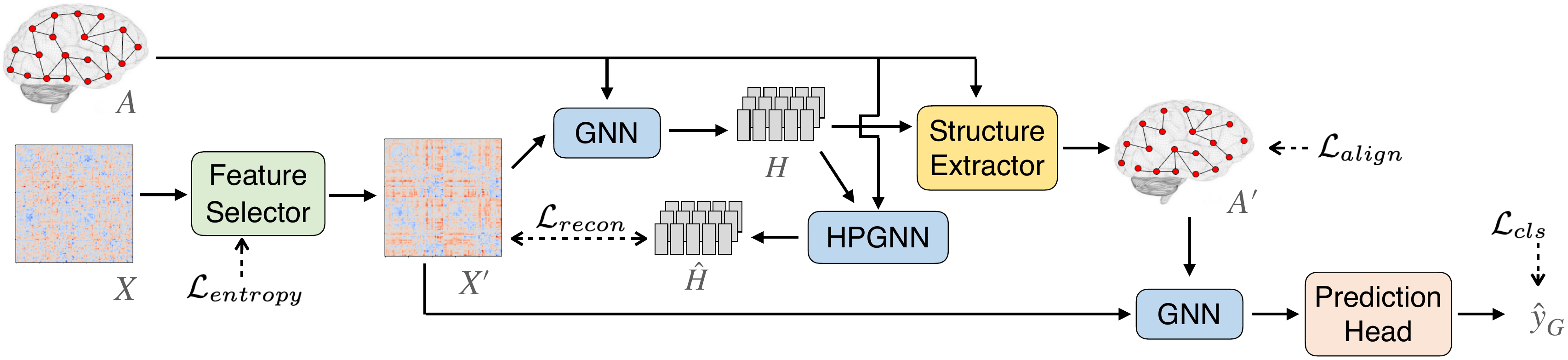}
\end{center}
\caption{The framework of BrainOOD.}
\label{fig:framework}
\end{figure*}

The aforementioned gap motivates us to propose a novel graph OOD architecture, called \textit{BrainOOD}, designed to offer both faithful interpretability and robust OOD generalizability. As shown in Figure \ref{fig:framework}, BrainOOD is composed of three main components: a feature selector, a structure extractor, and several auxiliary losses. 
These components work together to overcome the limitations of existing methods, ensuring that the model effectively captures discriminative connections while maintaining interpretability. The following sections provide a detailed description of each component and outline how they contribute to the promising performance and interpretability of BrainOOD.

\subsection{Feature Selection via Reconstruction}
Brain network data can contain noise in specific ROIs, where GNNs may even amplify the noise due to the smoothing nature in message passing.
This may further limit the extraction of useful information for the GIB objective.
To address this, we introduce a learnable masking mechanism that filters out irrelevant connections and focuses on the most informative node features. This is followed by a reconstruction loss to identify key distinguishing features.

Given an input brain network $G = (\boldsymbol{X}, \boldsymbol{A})$, the masked features are obtained as:
\begin{equation}
\label{eq:mask}
\small
\boldsymbol{X}^\prime = \boldsymbol{X} \odot \boldsymbol{M}, \quad
\boldsymbol{M} = \operatorname{Dropout} \left(\delta(\boldsymbol{W}_{mask} \boldsymbol{W}_{mask}^\mathsf{T}) \right),
\end{equation} 
where $\odot$ is the Hadamard product, $\boldsymbol{W}_{mask} \in \mathbb{R}^{n \times d}$ is the learnable mask embedding, and $\delta$ is the sigmoid function. We employ an entropy loss as a sparsity constraint, to compel the model to prioritize the most informative connections and prevent a smooth mask. The entropy loss is formulated as follows:
\begin{equation}
\label{eq:entropy_loss}
\small
\mathcal{L}_{entropy} = \frac{1}{n} \sum^n_{i=1} \operatorname{entropy} \left( \boldsymbol{M}(i,:) \right), \quad
\operatorname{entropy}(\boldsymbol{p}) = - \sum_{j=1}^n \boldsymbol{p}_j \log(\boldsymbol{p}_j).
\end{equation}
%
%
A GNN is subsequently employed to encode the brain network:
\begin{equation}
\label{eq:gnn}
\small
\boldsymbol{H} = \operatorname{GNN}(\boldsymbol{X}^\prime, \boldsymbol{A}).
\end{equation} 
It is well-known that GNN-based methods typically smooth node features across the graph, which can amplify noise in specific ROIs. To address this issue, we introduce a high-pass GNN to recover the input node features, guiding the model to learn the most informative features through a reconstruction loss:
\begin{equation}
\label{eq:recon_x}
\small
\hat{\boldsymbol{X}} = \operatorname{Tanh}(\hat{\boldsymbol{H}} \hat{\boldsymbol{H}^{\mathsf{T}}}) \odot \boldsymbol{M}, \quad
\hat{\boldsymbol{H}} = \operatorname{HPGNN}(\boldsymbol{H}, \boldsymbol{A}),
\end{equation}
\begin{equation}
\label{eq:recon_loss}
\small
\mathcal{L}_{recon} = \operatorname{MSE}(\hat{\boldsymbol{X}}, \boldsymbol{X}^\prime) = \frac{1}{n} \| \hat{\boldsymbol{X}} - \boldsymbol{X}^\prime \|_F^2,
\end{equation} 
where $\| \cdot \|_F $ denotes the Frobenius norm. Herein, $\operatorname{Tanh}(\cdot)$ serves to scale the range of the reconstructed features to align with the input connectivity matrix, while the self-multiplication operation is designed to ensure the output exhibits the symmetry property inherent in the connectivity matrix. This operation mimics the structure of the input data, making it easier for the model to capture meaningful patterns during reconstruction. By minimizing the Mean Square Error (MSE) between $\hat{\boldsymbol{X}}$ and $\boldsymbol{X}^\prime$, the feature selector is trained to extract the most informative features $\boldsymbol{X}^\prime$, ensuring the reconstruction is faithful to the input and improving the overall representation quality.

\subsection{Structure Extraction by Discrete Sampling}
Apart from node features, graph structure in brain networks also contains noise, which requires the model to extract critical substructures. When implementing the subgraph extractor $g_\phi$ in our improved GIB framework,
we adopt the sampling strategy proposed by \citet{cheninterpretable}. Specifically, an edge scorer is first applied to each edge in the input adjacency matrix based on the output of GNN encoder (Eq. (\ref{eq:gnn})) as:
\begin{equation}
\label{eq:edge_scorer}
\small
\alpha_{v,u} = \operatorname{scorer}([\boldsymbol{H}_v | \boldsymbol{H}_u]), \quad
\forall(v, u), \boldsymbol{A}_{v,u} = 1,
\end{equation} 
where $[\cdot|\cdot]$ is the concatenation function and $\operatorname{scorer}(\cdot)$ can be arbitrary attention functions such as a simple MLP with Gumbel-softmax \citep{maddison2022concrete}. Thus the probability $\gamma_{v,u}$ of edge $(v,u)$ for sampling is defined as:
\begin{equation}
\label{eq:prob}
\small
\gamma_{v,u} = \delta((\alpha_{v, u} + D)/\tau),
\end{equation} 
where $\tau$ is the temperature hyperparameter, $\delta(\cdot)$ is the sigmoid function, and $D = \log U - \log(1-U)$, with $U \sim \operatorname{Uniform}(0, 1)$. To sample the discrete subgraph, we sample from the Bernoulli distributions on edges independently by $\boldsymbol{A}^\prime_{v, u} \sim \operatorname{Bern}(\gamma_{v, u})$.
%
%

Finally the generated desired causal subgraph $G_C^* = (\boldsymbol{X}^\prime, \boldsymbol{A}^\prime)$ is used to learn the node representation by $\boldsymbol{H}^\prime = \operatorname{GNN}(\boldsymbol{X^\prime}, \boldsymbol{A}^\prime)$. This sampling is done for $k$ times to do the independent prediction and obtain the logits $\hat{y}_i$. The final prediction is computed by the average of the $k$ simulated predictions: $\hat{y}_G=\frac{1}{k} \sum_{i=1}^{k} \hat{y}_i$.

\subsection{Loss Functions}
\label{subsec:loss}

For brain networks, identifying specific ROIs or connections that correlate with neurological conditions is crucial for advancing our understanding of brain function and pathology. This task differs from traditional graph OOD methods, such as those proposed by \citet{wu2022dir}, \citet{miao2022interpretable}, and \citet{cheninterpretable}, which focus on extracting invariant substructures across different graphs. While such methods work well for general graph analysis, they fall short in brain network analysis, where the same structural patterns involving distinct ROIs can reflect varying functional roles in brain activities (as shown in Figure \ref{fig:example}). 

In BrainOOD, we aim to discover key discriminative connections, rather than merely identifying invariant substructures. These connections may hold vital clues to understanding conditions like Alzheimer's and Autism by revealing the functional relationships between brain regions. To address this, we propose an alignment loss that encourages the structure extractor to consistently select the same connections across all brain networks within a batch:
\begin{equation}
\label{eq:align_loss}
\small
\mathcal{L}_{align} = \frac{1}{n^2} \sum_{i=1}^{n} \sum_{j=1}^{n} \sigma_{i,j}^\prime,
\end{equation} 
where $\sigma^\prime$ is the standard deviation of all the $\boldsymbol{A}^\prime$ in the batch. By applying this constraint, BrainOOD identifies the most informative connections, promoting both generalizability and interpretability in brain network analysis.

To incorporate domain knowledge and facilitate model convergence during optimization, we utilize 4 loss functions to guide the end-to-end training. Specifically, (1) a commonly-used cross-entropy loss \citep{cox1958regression} $\mathcal{L}_{cls} = \operatorname{cross\_entropy}(\hat{y}_G, y_G)$  for graph classification; (2) an entropy loss $\mathcal{L}_{entropy}$ (Eq. (\ref{eq:entropy_loss})) for mask sparsification; (3) a reconstruction loss $\mathcal{L}_{recon}$ (Eq. (\ref{eq:recon_loss})) to enforce the GNN to encode the most discriminative information; (4) an alignment loss $\mathcal{L}_{align}$ (Eq. (\ref{eq:align_loss})) to constrain node-identity awareness. The total loss is computed by:
\begin{equation}
\label{eq:total_loss}
\small
\mathcal{L}_{total} = \mathcal{L}_{cls} + \lambda_1 * \mathcal{L}_{entropy} + \lambda_2 * \mathcal{L}_{recon} + \lambda_3 * \mathcal{L}_{align},
\end{equation} 
\noindent
where $\lambda_1$, $\lambda_2$ and $\lambda_3$ are trade-off hyperparameters.

\section{Experimental Results}
\label{sec:exp}

\subsection{Baseline Models}

We evaluate the proposed BrainOOD framework against a comprehensive set of baseline models, including \textbf{5 General OOD Methods}: ERM~\citep{goyal2017accurate}, Deep Coral~\citep{sun2016deep}, IRM~\citep{arjovsky2019invariant}, GroupDRO~\citep{sagawa2019distributionally} and VREx~\citep{krueger2021out}; \textbf{4 Graph OOD Methods}: Mixup~\citep{zhang2018mixup}, DIR~\citep{wu2022dir}, GSAT~\citep{miao2022interpretable} and GMT~\citep{cheninterpretable} (All these graph OOD methods and BrainOOD are incorporated with GIN backbone for fair comparison); \textbf{2 conventional machine learning methods}: Support Vector Machine (SVM) and Logistic Regression (LR) Classifier from scikit-learn~\citep{pedregosa2011scikit}, where these ML methods take the flattened upper-triangle connectivity matrix as vector input, instead of using the brain network; \textbf{3 General-Purpose GNNs}: GCN~\citep{kipf2016semi}, GIN~\citep{xu2018powerful} and GAT~\citep{velivckovic2017graph}; \textbf{4 Neural Networks Tailored for Brain Networks}:  BrainNetCNN~\citep{kawahara2017brainnetcnn}, BrainGNN~\citep{li2021braingnn}, ContrastPool~\citep{xu2024contrastive} and Contrasformer~\citep{xu2024contrasformer}.
The detailed baseline description and implementation of these experiments are provided in Appendices \ref{app:baseline} and \ref{app:imple_detail}, respectively.

\begin{table}[h]
\vspace{-0.15in}
\small
\centering
\caption{Results over 10-fold-CV (Average Accuracy ± Standard Deviation). The best result is highlighted in \textbf{bold} while the runner-up is highlighted in \underline{underline}.}
\setlength\tabcolsep{4pt}
\begin{tabular}{l|cc|cc|cc}
\hline
\multirow{2}{*}{OOD Model} & \multicolumn{2}{c|}{ABIDE}  & \multicolumn{2}{c|}{ADNI (5-class)}   & \multicolumn{2}{c}{ADNI (6-class)}    \\
                           & ID     & OOD    & ID      & OOD   & ID      & OOD  \\ \hline
GCN                        & 63.69 ± 3.20 & 56.45 ± 5.52 & 59.95 ± 8.20 & 55.32 ± 10.23 & 61.01 ± 9.53 & 59.16 ± 11.75\\
BrainNetCNN                        & \textbf{65.50} ± 4.77 & \underline{60.38} ± 7.07 & 62.08 ± 6.81 & 55.02 ± 11.10 & 61.26 ± 7.26 & 56.73 ± 11.69 \\ \hline
ERM                        & 59.17 ± 6.99 & 56.73 ± 5.99 & 60.86 ± 9.17 & 60.81 ± 13.47 & 60.55 ± 10.11 & 59.32 ± 15.67 \\
Deep Coral                 & 60.40 ± 5.34 & 56.95 ± 5.94 & 62.22 ± 8.25 & 60.39 ± 15.51 & 58.25 ± 10.26 & 57.28 ± 13.55 \\
IRM                        & 58.73 ± 7.07 & 57.34 ± 8.74 & 61.94 ± 9.13 & 60.89 ± 11.32 & \underline{62.36} ± 7.05 & 58.96 ± 13.42 \\
GroupDRO                   & 58.74 ± 8.43 & 58.83 ± 8.54 & 61.86 ± 8.34 & 57.34 ± 15.27 & 60.33 ± 7.74 & 54.33 ± 12.42 \\
VREx                       & 50.82 ± 2.11 & 52.08 ± 5.29 & 61.12 ± 6.71 & 55.64 ± 13.66 & 55.99 ± 9.45 & 50.07 ± 12.29 \\
Mixup                      & 62.06 ± 7.07 & 54.90 ± 7.71 & 62.82 ± 8.25 & 59.50 ± 12.81 & 60.51 ± 9.94 & 59.36 ± 13.73 \\
DIR                        & 59.77 ± 4.28 & 58.52 ± 9.61 & \underline{65.83} ± 9.49 & 57.99 ± 14.82 & 58.48 ± 7.70 & 58.19 ± 16.09\\
GSAT                       & 61.32 ± 6.37 & 57.57 ± 5.67 & 62.02 ± 8.77 & 60.27 ± 15.04 & 61.55 ± 10.55 & \underline{60.79} ± 13.97\\
GMT                        & 61.11 ± 6.30 & 59.73 ± 6.95 & 62.81 ± 6.54 & \underline{60.93} ± 13.27 & 61.65 ± 11.37 & 58.13 ± 13.53 \\
BrainOOD                   & \underline{64.07} ± 4.58 & \textbf{64.81} ± 9.01 & \textbf{66.09} ± 6.30 & \textbf{62.26} ± 15.83 & \textbf{65.71} ± 10.34 & \textbf{64.07} ± 14.99 \\ \hline
\end{tabular}
\label{tab:main_results}\vspace{-0.15in}
\end{table}

\begin{table}[h]
\setlength\tabcolsep{4pt}
\centering
\small
\caption{Results of more evaluation metrics over 10-fold-CV on the overall test set of ABIDE and ADNI datasets. The best result is highlighted in \textbf{bold} while the runner-up is highlighted in \underline{underline}. For multiclass dataset of ADNI, all other metrics are the same as accuracy.}
\scalebox{0.85}{
\begin{tabular}{c|ccccc|c|c}
\hline
\multirow{2}{*}{Model} & \multicolumn{5}{c|}{ABIDE}   & ADNI (5-class)  & ADNI (6-class)       \\
                       & Accuracy          & Precision     & Recall        & micro-F1           & ROC-AUC      & Accuracy          \\ \hline
SVM & 61.56 ± 4.04 & 61.10 ± 3.57 & 63.02 ± 3.57 & 61.53 ± 7.28 & 60.89 ± 4.31 & 62.88 ± 4.75 & 61.24 ± 5.53 \\
LR & 61.23 ± 3.93 & 63.16 ± 2.89 & 62.72 ± 6.45 & 62.77 ± 3.81 & 61.32 ± 2.93 & 61.58 ± 4.52 & 61.05 ± 6.11 \\\hline
GCN                    & 61.85 ± 4.39 & 60.13 ± 3.94  & 58.45 ± 10.67 & 58.88 ± 7.06 & 61.71 ± 4.59 & 61.92 ± 9.53 & 60.92 ± 4.13 \\
GIN                    & 56.49 ± 3.40 & \underline{62.78} ± 12.71 & 28.52 ± 10.69 & 37.46 ± 7.56 & 55.22 ± 3.23 & 58.78 ± 9.53 & 59.29 ± 3.72 \\
GAT                    & 63.12 ± 4.72 & 61.50 ± 5.22  & 61.29 ± 6.75  & 61.20 ± 5.06 & 63.07 ± 4.67 & 60.94 ± 6.58 & 60.07 ± 5.34 \\ \hline
BrainNetCNN            & \underline{63.80} ± 4.44 & 62.38 ± 6.11  & 63.34 ± 8.11  & 62.35 ± 4.63 & \textbf{63.79} ± 4.32 & 59.77 ± 8.69 & 58.76 ± 3.09 \\
BrainGNN                  & 60.00 ± 3.96 & 58.94 ± 4.98  & 54.34 ± 7.30  & 56.23 ± 4.89 & 59.76 ± 3.93 & 62.08 ± 8.93 & 62.40 ± 4.44 \\
ContrastPool           & 62.00 ± 2.97 & 56.02 ± 3.92  & \textbf{68.46} ± 12.60 & 62.84 ± 5.69 & 62.57 ± 3.93 & 61.22 ± 1.87 & 60.00 ± 5.54 \\
Contrasformer          & 63.53 ± 3.03 & 60.73 ± 3.23  & \underline{65.87} ± 6.30  & \underline{63.01} ± 3.43 & \underline{63.67} ± 3.02 & \underline{63.52} ± 3.10 & \underline{63.58} ± 6.06 \\ \hline
ERM                    & 60.00 ± 3.35 & 57.84 ± 5.15  & 57.43 ± 4.78  & 56.88 ± 4.99 & 57.47 ± 4.67 & 60.69 ± 4.32& 59.60 ± 5.04 \\
Deep Coral             & 59.71 ± 4.55 & 60.50 ± 5.08  & 59.46 ± 5.21  & 58.22 ± 5.54 & 58.97 ± 4.89 & 61.47 ± 3.42 & 57.74 ± 6.43 \\
IRM                    & 60.15 ± 4.97 & 61.34 ± 5.23  & 59.84 ± 4.61  & 58.81 ± 5.01 & 59.89 ± 4.56 & 61.16 ± 4.69 & 60.93 ± 4.96 \\
GroupDRO               & 59.70 ± 2.89 & 60.91 ± 3.16  & 59.17 ± 3.59  & 58.24 ± 3.34 & 59.65 ± 3.13 & 59.84 ± 4.92 & 57.60 ± 4.16 \\
VREx                   & 57.47 ± 4.64 & 59.36 ± 4.92  & 53.82 ± 9.28  & 54.15 ± 7.44 & 57.06 ± 4.81 & 58.76 ± 3.79 & 54.11 ± 5.54 \\
Mixup                  & 60.30 ± 3.28 & 59.43 ± 5.00  & 58.16 ± 4.53  & 56.95 ± 3.98 & 58.23 ± 4.38 & 61.08 ± 3.27& 60.00 ± 3.89 \\
DIR                    & 59.27 ± 6.41 & 60.45 ± 6.76  & 59.48 ± 6.84  & 58.22 ± 6.78 & 59.35 ± 6.76 & 62.16 ± 4.82 & 58.13 ± 6.29 \\
GSAT                   & 59.38 ± 3.54 & 59.73 ± 4.62  & 59.11 ± 4.28  & 58.15 ± 4.22 & 58.76 ± 4.01 & 60.92 ± 7.30 & 61.00 ± 6.02 \\
GMT                    & 60.95 ± 3.50 & 60.32 ± 3.29  & 59.96 ± 3.59  & 59.41 ± 3.69 & 59.81 ± 3.52 & 61.61 ± 6.44 & 60.00 ± 5.80 \\
BrainOOD (ours)              & \textbf{63.95} ± 4.65 & \textbf{65.72} ± 5.24  & 63.37 ± 4.29  & \textbf{63.42} ± 4.86 & 63.52 ± 4.28 & \textbf{64.18} ± 5.48 & \textbf{64.80} ± 5.36 \\ \hline
\end{tabular}
}
\label{tab:more_metrics}
\end{table}

\subsection{Main Results}

We first compare BrainOOD with existing baselines in terms of in-domain (ID) and OOD classification accuracy. The results on 2 brain network datasets over 10-fold cross-validation (CV) are reported in Table \ref{tab:main_results}. Apart from classifying ADNI into 6 classes, we also conduct experiment by a 5-class setting by merging MCI with LMCI to align with most AD diagnosis/prognosis studies in the literature. This is due to the MCI defined in ADNI 1 corresponding to LMCI in ANDI GO/2. Although non-OOD methods (GCN and BrainNetCNN) achieve good accuracy on ID set, they failed to generalize to OOD data. Most OOD algorithms have comparable performance with ERM, showing the difficulty of achieving invariant prediction in brain networks. While these graph OOD methods (Mixup, DIR, GSAT and GMT) apply well to graph topology, their failure to consider the unique characteristics of brain networks creates a performance bottleneck.
On the contrary, our proposed BrainOOD leads to non-trivial improvements on both ID and OOD sets for all datasets. Especially, for the performance on OOD set, the improvement is up to 7.34\% ((64.81\% - 60.38\%) / 60.38\% = 7.34\% compared with BrainNetCNN). We further provide a deeper analysis for the performance distribution of graph OOD methods on each fold in Appendix \ref{app:ood_acc}. BrainOOD consistently achieves top performance across multiple folds and maintains robustness in worst-case scenarios, demonstrating strong generalization capabilities to unseen sites.

\begin{figure}[h]
\centering
\includegraphics[width=0.4\textwidth]{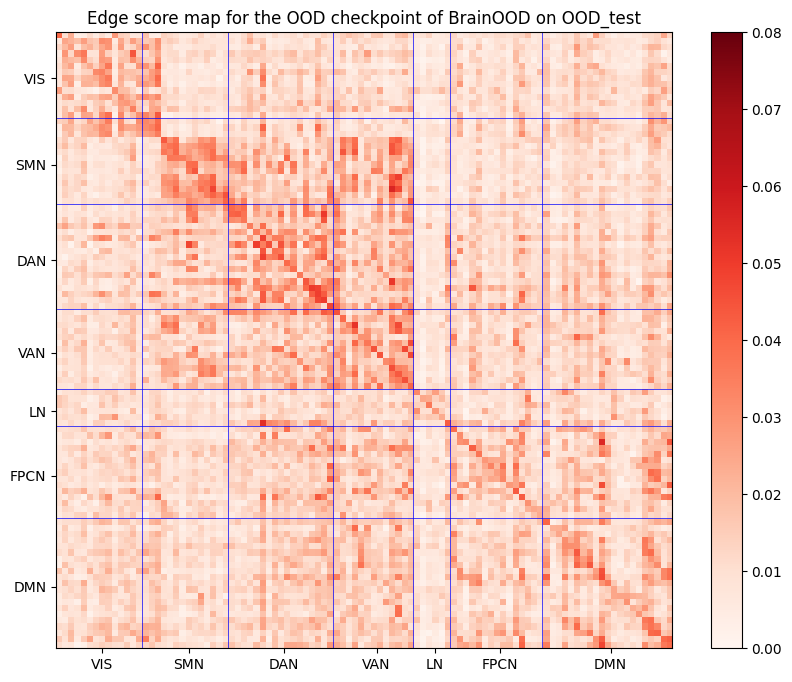}
\includegraphics[width=0.4\textwidth]{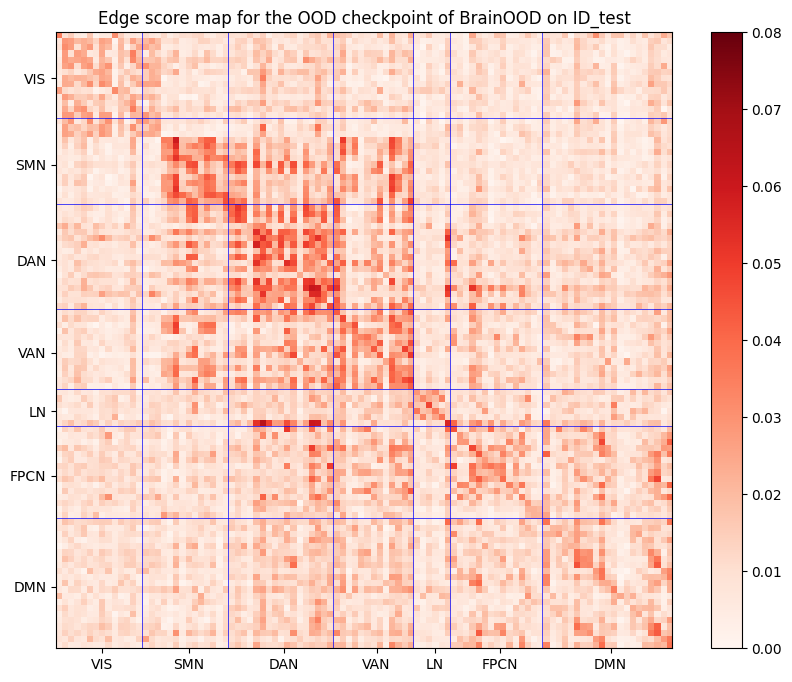}
\includegraphics[width=0.4\textwidth]{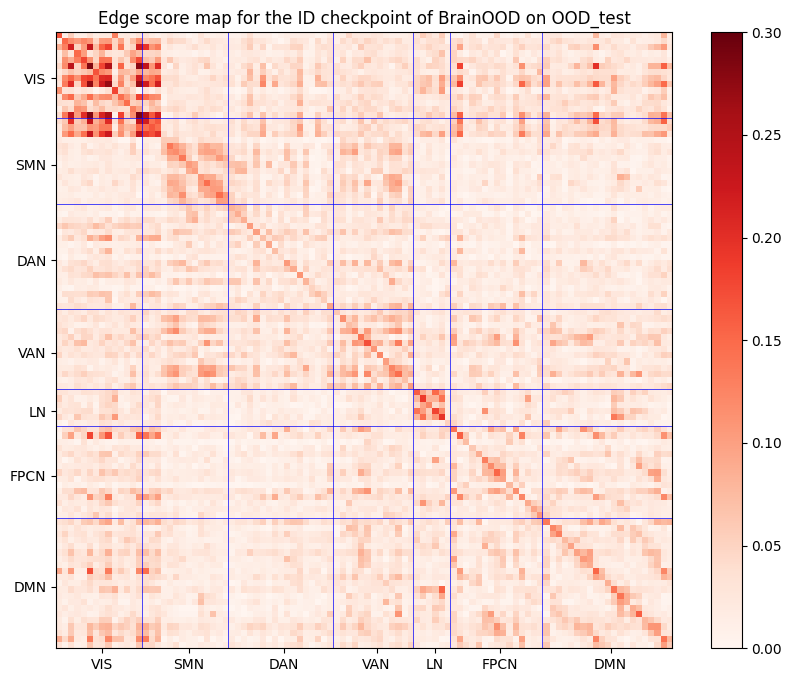}
\includegraphics[width=0.4\textwidth]{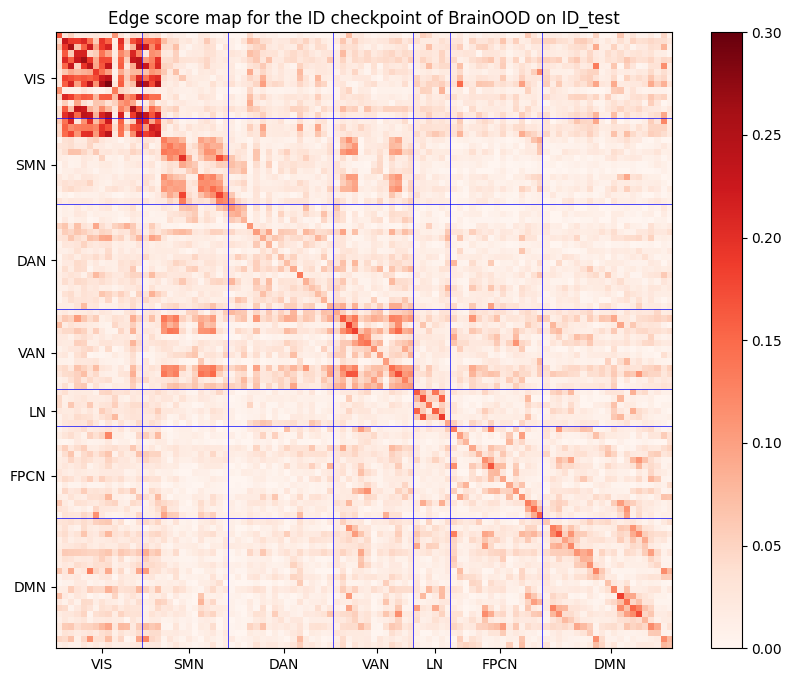}
\caption{Edge score map visualization for ID/OOD checkpoints on ID/OOD test set of ABIDE dataset. VIS = visual network; SMN = somatomotor network; DAN = dorsal attention network; VAN = ventral attention network; LN = limbic network; FPCN = frontoparietal control network; DMN = default mode network.}
\label{fig:vis_abide_brainood}
\end{figure}

To further compare BrainOOD with other general-purpose GNNs and neural networks specifically designed for brain networks, we report the results on the overall test sets in Table \ref{tab:more_metrics}. Our proposed BrainOOD emerges as clear winner across both datasets. Interestingly, all existing OOD methods perform poorly, struggling even to surpass the simple GNN baselines. This suggests that current approaches to extracting invariant subgraphs are ineffective for brain networks and highlights the need for OOD algorithms that account for the unique characteristics of brain data. Notably, compared with the GIN backbone, incorporating our proposed OOD framework yields a significant 12.2\% improvement, further verifying the effectiveness and necessity of BrainOOD in brain network analysis.

\subsection{Model Interpretation}

In the domain of neurodegenerative disorder diagnosis, identifying significant ROIs and connections associated with predictions is critical, as these serve as potential biomarkers for diseases. For this study, we leverage edge scores from the structure extractor in BrainOOD to generate heat maps, providing interpretability of the model's predictions. These score maps are visualized using the Nilearn toolbox~\citep{abraham2014machine}.  Figure \ref{fig:vis_abide_brainood} shows score maps for both ID and OOD checkpoints on the respective test sets from the ABIDE dataset, where higher scores signify stronger classification potential for ASD. We assessed the connections highlighted by our model in relation to Yeo's 7 networks~\citep{yeo2011organization} that may be linked to the disorder. As shown in Figure \ref{fig:vis_abide_brainood}, the score maps for the same checkpoint are consistent across both the ID and OOD test sets, suggesting that the model captures invariant patterns relevant to OOD subjects. Additionally, comparing different checkpoints on the same test sets reveals that both ID and OOD checkpoints identify common connections within key networks such as the somatomotor network (SMN), ventral attention network (VAN), and limbic network (LN), which are often associated with ASD~\citep{hong2019atypical,farrant2016atypical}. Interestingly, the score maps from the ID checkpoints tend to be sparser compared to those from OOD checkpoints. Furthermore, some connections are uniquely highlighted at different checkpoints, such as those within the visual network (VIS) for the ID checkpoint and within the dorsal attention network (DAN) for the OOD checkpoint.

\begin{wrapfigure}{r}{0.4\textwidth}
\vspace{-0.1in}
\begin{center}
\includegraphics[width=\linewidth]{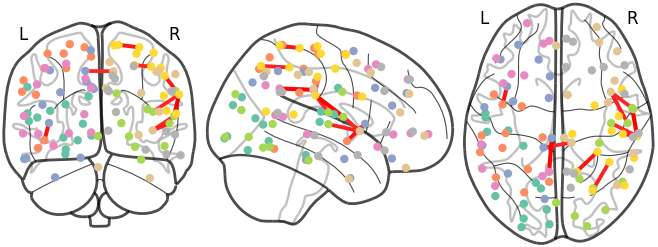}
\end{center}
\caption{The visualization of the top 10 connections with the highest score on ABIDE OOD set.}
\label{fig:brain_abide}
\vspace{-0.1in}
\end{wrapfigure}

To pinpoint the connections most significant for the causal subgraph, we selected the top 10 connections with the highest scores. Figure \ref{fig:brain_abide} highlights connections between the posterior, temporal occipital parietal regions in the ABIDE dataset, suggesting potential ASD-specific neural mechanisms. These regions align with prior research, which has identified them as critical areas in ASD studies~\citep{ciaramidaro2018transdiagnostic}. Notably, these findings resonate with the discovery that adolescents with ASD exhibit hypo-activation in key visuoperceptual regions, particularly in the right hemisphere, as well as in affective and motivational face-processing areas~\citep{scherf2015individual}. A discussion of AD findings from the ADNI dataset is provided in Appendix \ref{app:case_adni}.

\subsection{Ablation Study}

To verify the effectiveness of our proposed components in BrainOOD, we test our design of the loss functions by disabling them one by one. 
The results are reported in Table \ref{tab:ablation}, where ``feat'' and ``adj'' represent what we used as the feature matrix and adjacency matrix for the final prediction, respectively. We can observe that all of the auxiliary losses and components are effective in boosting the model performance. Besides, we find that the reconstruction loss and the alignment loss are important to ensure the ability of BrainOOD to generalize to the OOD set. This observation indicates the necessity of selecting information on both the feature and structure levels. We include the detailed hyperparameter sensitivity analysis in Appendix \ref{app:hyper}.

\begin{table}[h]\vspace{-0.15in}
\centering
\small
\caption{Ablation study on important components in BrainOOD on ABIDE dataset.}
\label{tab:ablation}
\begin{tabular}{cccccccc}
\hline
feat & adj & $\mathcal{L}_{entorpy}$ & $\mathcal{L}_{recon}$ & $\mathcal{L}_{align}$ & ID acc         & OOD acc        & overall acc        \\ \hline
 $\boldsymbol{X}^\prime$  &  $\boldsymbol{A}^\prime$ &       & \checkmark        & \checkmark        & 63.92 ± 4.13   & 63.70 ± 4.53   & 63.12 ± 2.50   \\
$\boldsymbol{X}^\prime$  &  $\boldsymbol{A}^\prime$ &  \checkmark          &          & \checkmark        & 62.82   ± 4.19 & 61.37   ± 7.13 & 61.85   ± 4.53 \\
$\boldsymbol{X}^\prime$  &  $\boldsymbol{A}^\prime$ &  \checkmark          & \checkmark        &          & 63.26   ± 3.44 & 60.43   ± 5.45 & 61.85   ± 2.83 \\
$\boldsymbol{X}$  &  $\boldsymbol{A}^\prime$ &  \checkmark          & \checkmark        & \checkmark        & 63.56 ± 4.40   & 62.26 ± 5.68   & 62.69 ± 3.42 \\
$\boldsymbol{X}^\prime$  &  $\boldsymbol{A}$ &  \checkmark          & \checkmark        & \checkmark        & 63.71 ± 5.97  & 55.40 ± 8.95   & 60.10 ± 3.47 \\
$\boldsymbol{X}^\prime$  &  $\boldsymbol{A}^\prime$ &  \checkmark          & \checkmark        & \checkmark        & \textbf{64.07} ± 4.58   & \textbf{64.81} ± 9.01   & \textbf{63.95} ± 4.65 \\\hline
\end{tabular}\vspace{-0.15in}
\end{table}

\section{Related Works}


OOD or distribution shift is a longstanding problem in machine learning \citep{goyal2017accurate,zhang2018mixup,sagawa2019distributionally,krueger2021out}. Most existing graph OOD methods aim to extract invariant subgraphs across all samples to enhance model generalization under distribution shifts. GIL \citep{li2022learning} is a pioneering GNN-based model that identifies invariant subgraphs for graph classification tasks. It explores invariant graph representation learning in mixed latent environments without requiring labeled environments. DIR \citep{wu2022dir} introduces a causal inference approach to identify invariant causal parts through causal interventions. However, DIR involves a complex iterative process of breaking and assembling subgraphs during training. A more straightforward approach is GSAT \citep{miao2022interpretable}, which is based on the information bottleneck principle and learns invariant subgraphs by reducing attention stochasticity. RGCL \citep{li2022let} combines invariant rationale discovery with contrastive learning to improve both generalization and interpretability. CIGA \citep{chen2022learning} proposes an information-theoretic objective to extract invariant subgraphs, offering a theoretical guarantee for handling distribution shifts under different Structural Causal Models, which inspired a number of follow-up approaches~\citep{gala,yao2024empowering}. Similarly, GMT \citep{cheninterpretable} focuses on extracting interpretable subgraphs by accurately approximating subgraph multilinear extensions, ensuring both interpretability and generalization under OOD conditions.  A common finding across these invariant learning-based methods is the dependence on the diversity of environments. To address this, IGM \citep{jia2024graph} introduces a co-mixup strategy that combines environment and invariant mixups to generate diverse environments. These OOD methods that focus on extracting causal subgraphs work well in molecular and social networks but face challenges in brain network analysis due to the unique noise in both structures and features. These methods often overlook the selection of important node features, reducing their effectiveness for brain networks. Additionally, invariant subgraphs identified by these methods may not adequately capture the distinct functional implications of different brain regions, underscoring the need for a specialized approach. We include the discussion of more related works about brain network analysis with GNNs in Appendix \ref{app:related}.


\section{Conclusion}
In this work, we introduced BrainOOD, a novel framework designed to tackle the dual challenges of OOD generalization and interpretability in brain network analysis. BrainOOD improves the representation power of GNNs through a feature selection process and a learnable masking mechanism, addressing the unique characteristics of brain networks by focusing on identifying critical connections rather than invariant substructures. The model's reconstruction loss further enhances its ability to reveal causally interpretable brain regions. Our extensive evaluations across 16 existing methods demonstrate that BrainOOD significantly outperforms both general-purpose and brain-specific GNNs, achieving up to an 8.5\% improvement over existing graph OOD methods. Importantly, the model not only enhances OOD generalization but also extracts scientifically meaningful patterns that align with established knowledge in neuroscience. By presenting the first OOD benchmark dataset for brain network analysis, we provide a valuable resource for future research in improving both the generalizability and interpretability of models in this important domain of scientific research.


\subsubsection*{Acknowledgments}

We thank the reviewers for their valuable comments. This research/project is supported by the Ministry of Education, Singapore under its MOE Academic Research Fund Tier 2 (STEM RIE2025 Award MOE-T2EP20220-0006) and Tier 1 (RG16/24). Any opinions, findings and conclusions or recommendations expressed in this material are those of the author(s) and do not reflect the views of the Ministry of Education, Singapore. YQ and JC are supported by Research Grants 8601116, 8601594, and 8601625 from the UGC of Hong Kong.

\bibliography{iclr2025_conference}

\begin{thebibliography}{63}
\providecommand{\natexlab}[1]{#1}
\providecommand{\url}[1]{\texttt{#1}}
\expandafter\ifx\csname urlstyle\endcsname\relax
  \providecommand{\doi}[1]{doi: #1}\else
  \providecommand{\doi}{doi: \begingroup \urlstyle{rm}\Url}\fi

\bibitem[Abraham et~al.(2014)Abraham, Pedregosa, Eickenberg, Gervais, Mueller, Kossaifi, Gramfort, Thirion, and Varoquaux]{abraham2014machine}
Alexandre Abraham, Fabian Pedregosa, Michael Eickenberg, Philippe Gervais, Andreas Mueller, Jean Kossaifi, Alexandre Gramfort, Bertrand Thirion, and Ga{\"e}l Varoquaux.
\newblock Machine learning for neuroimaging with scikit-learn.
\newblock \emph{Frontiers in neuroinformatics}, 8:\penalty0 14, 2014.

\bibitem[Arjovsky et~al.(2019)Arjovsky, Bottou, Gulrajani, and Lopez-Paz]{arjovsky2019invariant}
Martin Arjovsky, L{\'e}on Bottou, Ishaan Gulrajani, and David Lopez-Paz.
\newblock Invariant risk minimization.
\newblock \emph{arXiv preprint arXiv:1907.02893}, 2019.

\bibitem[Boyle et~al.(2024)Boyle, Klinger, Shirzadi, Coughlan, Seto, Properzi, Townsend, Yuan, Scanlon, Jutten, et~al.]{boyle2024left}
Rory Boyle, HM~Klinger, Z~Shirzadi, GT~Coughlan, M~Seto, MJ~Properzi, Diana~L Townsend, Ziwen Yuan, C~Scanlon, Roos~J Jutten, et~al.
\newblock Left frontoparietal control network connectivity moderates the effect of amyloid on cognitive decline in preclinical alzheimer’s disease: The a4 study.
\newblock \emph{The Journal of Prevention of Alzheimer's Disease}, 11\penalty0 (4):\penalty0 881--888, 2024.

\bibitem[Chan et~al.(2022)Chan, Yew, and Rajapakse]{chan2022semi}
Yi~Hao Chan, Wei~Chee Yew, and Jagath~C Rajapakse.
\newblock Semi-supervised learning with data harmonisation for biomarker discovery from resting state fmri.
\newblock In \emph{International Conference on Medical Image Computing and Computer-Assisted Intervention}, pp.\  441--451. Springer, 2022.

\bibitem[Chen et~al.(2022{\natexlab{a}})Chen, Zhang, Bian, Yang, Kaili, Xie, Liu, Han, and Cheng]{chen2022learning}
Yongqiang Chen, Yonggang Zhang, Yatao Bian, Han Yang, MA~Kaili, Binghui Xie, Tongliang Liu, Bo~Han, and James Cheng.
\newblock Learning causally invariant representations for out-of-distribution generalization on graphs.
\newblock \emph{Advances in Neural Information Processing Systems}, 35:\penalty0 22131--22148, 2022{\natexlab{a}}.

\bibitem[Chen et~al.(2022{\natexlab{b}})Chen, Zhang, Bian, Yang, Ma, Xie, Liu, Han, and Cheng]{ciga}
Yongqiang Chen, Yonggang Zhang, Yatao Bian, Han Yang, Kaili Ma, Binghui Xie, Tongliang Liu, Bo~Han, and James Cheng.
\newblock Learning causally invariant representations for out-of-distribution generalization on graphs.
\newblock In \emph{Advances in Neural Information Processing Systems}, 2022{\natexlab{b}}.

\bibitem[Chen et~al.(2023)Chen, Bian, Zhou, Xie, Han, and Cheng]{gala}
Yongqiang Chen, Yatao Bian, Kaiwen Zhou, Binghui Xie, Bo~Han, and James Cheng.
\newblock Does invariant graph learning via environment augmentation learn invariance?
\newblock In \emph{Advances in Neural Information Processing Systems}, 2023.

\bibitem[Chen et~al.(2024)Chen, Bian, Han, and Cheng]{cheninterpretable}
Yongqiang Chen, Yatao Bian, Bo~Han, and James Cheng.
\newblock How interpretable are interpretable graph neural networks?
\newblock In \emph{Forty-first International Conference on Machine Learning}, 2024.

\bibitem[Ciaramidaro et~al.(2018)Ciaramidaro, B{\"o}lte, Schlitt, Hainz, Poustka, Weber, Freitag, and Walter]{ciaramidaro2018transdiagnostic}
Angela Ciaramidaro, Sven B{\"o}lte, Sabine Schlitt, Daniela Hainz, Fritz Poustka, Bernhard Weber, Christine Freitag, and Henrik Walter.
\newblock Transdiagnostic deviant facial recognition for implicit negative emotion in autism and schizophrenia.
\newblock \emph{European Neuropsychopharmacology}, 28\penalty0 (2):\penalty0 264--275, 2018.

\bibitem[Cox(1958)]{cox1958regression}
David~R Cox.
\newblock The regression analysis of binary sequences.
\newblock \emph{Journal of the Royal Statistical Society: Series B (Methodological)}, 20\penalty0 (2):\penalty0 215--232, 1958.

\bibitem[Craddock et~al.(2013)Craddock, Benhajali, Chu, Chouinard, Evans, Jakab, Khundrakpam, Lewis, Li, Milham, et~al.]{craddock2013neuro}
Cameron Craddock, Yassine Benhajali, Carlton Chu, Francois Chouinard, Alan Evans, Andr{\'a}s Jakab, Budhachandra~Singh Khundrakpam, John~David Lewis, Qingyang Li, Michael Milham, et~al.
\newblock The neuro bureau preprocessing initiative: open sharing of preprocessed neuroimaging data and derivatives.
\newblock \emph{Frontiers in Neuroinformatics}, 7:\penalty0 27, 2013.

\bibitem[Dadi et~al.(2019)Dadi, Rahim, Abraham, Chyzhyk, Milham, Thirion, Varoquaux, Initiative, et~al.]{dadi2019benchmarking}
Kamalaker Dadi, Mehdi Rahim, Alexandre Abraham, Darya Chyzhyk, Michael Milham, Bertrand Thirion, Ga{\"e}l Varoquaux, Alzheimer's Disease~Neuroimaging Initiative, et~al.
\newblock Benchmarking functional connectome-based predictive models for resting-state fmri.
\newblock \emph{NeuroImage}, 192:\penalty0 115--134, 2019.

\bibitem[Farrant \& Uddin(2016)Farrant and Uddin]{farrant2016atypical}
Kristafor Farrant and Lucina~Q Uddin.
\newblock Atypical developmental of dorsal and ventral attention networks in autism.
\newblock \emph{Developmental science}, 19\penalty0 (4):\penalty0 550--563, 2016.

\bibitem[Fey \& Lenssen(2019)Fey and Lenssen]{Fey/Lenssen/2019}
Matthias Fey and Jan~E. Lenssen.
\newblock Fast graph representation learning with {PyTorch Geometric}.
\newblock In \emph{ICLR Workshop on Representation Learning on Graphs and Manifolds}, 2019.

\bibitem[Goyal(2017)]{goyal2017accurate}
P~Goyal.
\newblock Accurate, large minibatch sg d: training imagenet in 1 hour.
\newblock \emph{arXiv preprint arXiv:1706.02677}, 2017.

\bibitem[Guan et~al.(2021)Guan, Liu, Yang, Yap, Shen, and Liu]{guan2021multi}
Hao Guan, Yunbi Liu, Erkun Yang, Pew-Thian Yap, Dinggang Shen, and Mingxia Liu.
\newblock Multi-site mri harmonization via attention-guided deep domain adaptation for brain disorder identification.
\newblock \emph{Medical image analysis}, 71:\penalty0 102076, 2021.

\bibitem[Gui et~al.(2022)Gui, Li, Wang, and Ji]{gui2022good}
Shurui Gui, Xiner Li, Limei Wang, and Shuiwang Ji.
\newblock {GOOD}: A graph out-of-distribution benchmark.
\newblock In \emph{Thirty-sixth Conference on Neural Information Processing Systems Datasets and Benchmarks Track}, 2022.
\newblock URL \url{https://openreview.net/forum?id=8hHg-zs_p-h}.

\bibitem[Hong et~al.(2019)Hong, Vos~de Wael, Bethlehem, Lariviere, Paquola, Valk, Milham, Di~Martino, Margulies, Smallwood, et~al.]{hong2019atypical}
Seok-Jun Hong, Reinder Vos~de Wael, Richard~AI Bethlehem, Sara Lariviere, Casey Paquola, Sofie~L Valk, Michael~P Milham, Adriana Di~Martino, Daniel~S Margulies, Jonathan Smallwood, et~al.
\newblock Atypical functional connectome hierarchy in autism.
\newblock \emph{Nature communications}, 10\penalty0 (1):\penalty0 1022, 2019.

\bibitem[Ioffe \& Szegedy(2015)Ioffe and Szegedy]{10.5555/3045118.3045167}
Sergey Ioffe and Christian Szegedy.
\newblock Batch normalization: accelerating deep network training by reducing internal covariate shift.
\newblock In \emph{Proceedings of the 32nd International Conference on International Conference on Machine Learning - Volume 37}, ICML'15, pp.\  448–456. JMLR.org, 2015.

\bibitem[Jia et~al.(2024)Jia, Li, Yang, Tao, and Shi]{jia2024graph}
Tianrui Jia, Haoyang Li, Cheng Yang, Tao Tao, and Chuan Shi.
\newblock Graph invariant learning with subgraph co-mixup for out-of-distribution generalization.
\newblock In \emph{Proceedings of the AAAI Conference on Artificial Intelligence}, volume~38, pp.\  8562--8570, 2024.

\bibitem[Jiang et~al.(2020)Jiang, Liu, Crawford, Kochan, Brodaty, Sachdev, and Wen]{jiang2020stronger}
Jiyang Jiang, Tao Liu, John~D Crawford, Nicole~A Kochan, Henry Brodaty, Perminder~S Sachdev, and Wei Wen.
\newblock Stronger bilateral functional connectivity of the frontoparietal control network in near-centenarians and centenarians without dementia.
\newblock \emph{Neuroimage}, 215:\penalty0 116855, 2020.

\bibitem[Kawahara et~al.(2017)Kawahara, Brown, Miller, Booth, Chau, Grunau, Zwicker, and Hamarneh]{kawahara2017brainnetcnn}
Jeremy Kawahara, Colin~J Brown, Steven~P Miller, Brian~G Booth, Vann Chau, Ruth~E Grunau, Jill~G Zwicker, and Ghassan Hamarneh.
\newblock Brainnetcnn: Convolutional neural networks for brain networks; towards predicting neurodevelopment.
\newblock \emph{NeuroImage}, 146:\penalty0 1038--1049, 2017.

\bibitem[Kingma \& Ba(2014)Kingma and Ba]{kingma2014adam}
Diederik~P Kingma and Jimmy Ba.
\newblock Adam: A method for stochastic optimization.
\newblock \emph{arXiv preprint arXiv:1412.6980}, 2014.

\bibitem[Kipf \& Welling(2016)Kipf and Welling]{kipf2016semi}
Thomas~N Kipf and Max Welling.
\newblock Semi-supervised classification with graph convolutional networks.
\newblock \emph{arXiv preprint arXiv:1609.02907}, 2016.

\bibitem[Krueger et~al.(2021)Krueger, Caballero, Jacobsen, Zhang, Binas, Zhang, Le~Priol, and Courville]{krueger2021out}
David Krueger, Ethan Caballero, Joern-Henrik Jacobsen, Amy Zhang, Jonathan Binas, Dinghuai Zhang, Remi Le~Priol, and Aaron Courville.
\newblock Out-of-distribution generalization via risk extrapolation (rex).
\newblock In \emph{International conference on machine learning}, pp.\  5815--5826. PMLR, 2021.

\bibitem[Ktena et~al.(2017)Ktena, Parisot, Ferrante, Rajchl, Lee, Glocker, and Rueckert]{ktena2017distance}
Sofia~Ira Ktena, Sarah Parisot, Enzo Ferrante, Martin Rajchl, Matthew Lee, Ben Glocker, and Daniel Rueckert.
\newblock Distance metric learning using graph convolutional networks: Application to functional brain networks.
\newblock In \emph{Medical Image Computing and Computer Assisted Intervention- MICCAI 2017: 20th International Conference, Quebec City, QC, Canada, September 11-13, 2017, Proceedings, Part I 20}, pp.\  469--477. Springer, 2017.

\bibitem[Lanciano et~al.(2020)Lanciano, Bonchi, and Gionis]{lanciano2020explainable}
Tommaso Lanciano, Francesco Bonchi, and Aristides Gionis.
\newblock Explainable classification of brain networks via contrast subgraphs.
\newblock In \emph{Proceedings of the 26th ACM SIGKDD International Conference on Knowledge Discovery \& Data Mining}, pp.\  3308--3318, 2020.

\bibitem[Lei et~al.(2023)Lei, Zhu, Liang, Yang, Chen, Hu, Xie, Wei, Hao, Song, Wang, Xiao, Wang, and Han]{10198494}
Baiying Lei, Yun Zhu, Enmin Liang, Peng Yang, Shaobin Chen, Huoyou Hu, Haoran Xie, Ziyi Wei, Fei Hao, Xuegang Song, Tianfu Wang, Xiaohua Xiao, Shuqiang Wang, and Hongbin Han.
\newblock Federated domain adaptation via transformer for multi-site alzheimer’s disease diagnosis.
\newblock \emph{IEEE Transactions on Medical Imaging}, 42\penalty0 (12):\penalty0 3651--3664, 2023.
\newblock \doi{10.1109/TMI.2023.3300725}.

\bibitem[Li et~al.(2022{\natexlab{a}})Li, Zhang, Wang, and Zhu]{li2022learning}
Haoyang Li, Ziwei Zhang, Xin Wang, and Wenwu Zhu.
\newblock Learning invariant graph representations for out-of-distribution generalization.
\newblock \emph{Advances in Neural Information Processing Systems}, 35:\penalty0 11828--11841, 2022{\natexlab{a}}.

\bibitem[Li et~al.(2022{\natexlab{b}})Li, Wang, Zhang, Wu, He, and Chua]{li2022let}
Sihang Li, Xiang Wang, An~Zhang, Yingxin Wu, Xiangnan He, and Tat-Seng Chua.
\newblock Let invariant rationale discovery inspire graph contrastive learning.
\newblock In \emph{International conference on machine learning}, pp.\  13052--13065. PMLR, 2022{\natexlab{b}}.

\bibitem[Li et~al.(2019)Li, Dvornek, Zhou, Zhuang, Ventola, and Duncan]{li2019graph}
Xiaoxiao Li, Nicha~C Dvornek, Yuan Zhou, Juntang Zhuang, Pamela Ventola, and James~S Duncan.
\newblock Graph neural network for interpreting task-fmri biomarkers.
\newblock In \emph{Medical Image Computing and Computer Assisted Intervention--MICCAI 2019: 22nd International Conference, Shenzhen, China, October 13--17, 2019, Proceedings, Part V 22}, pp.\  485--493. Springer, 2019.

\bibitem[Li et~al.(2020)Li, Zhou, Dvornek, Zhang, Zhuang, Ventola, and Duncan]{li2020pooling}
Xiaoxiao Li, Yuan Zhou, Nicha~C Dvornek, Muhan Zhang, Juntang Zhuang, Pamela Ventola, and James~S Duncan.
\newblock Pooling regularized graph neural network for fmri biomarker analysis.
\newblock In \emph{Medical Image Computing and Computer Assisted Intervention--MICCAI 2020: 23rd International Conference, Lima, Peru, October 4--8, 2020, Proceedings, Part VII 23}, pp.\  625--635. Springer, 2020.

\bibitem[Li et~al.(2021)Li, Zhou, Dvornek, Zhang, Gao, Zhuang, Scheinost, Staib, Ventola, and Duncan]{li2021braingnn}
Xiaoxiao Li, Yuan Zhou, Nicha Dvornek, Muhan Zhang, Siyuan Gao, Juntang Zhuang, Dustin Scheinost, Lawrence~H Staib, Pamela Ventola, and James~S Duncan.
\newblock Braingnn: Interpretable brain graph neural network for fmri analysis.
\newblock \emph{Medical Image Analysis}, 74:\penalty0 102233, 2021.

\bibitem[Liu et~al.(2023)Liu, Wu, Li, Liu, Tian, and Huang]{10012442}
Xingdan Liu, Jiacheng Wu, Wenqi Li, Qian Liu, Lixia Tian, and Huifang Huang.
\newblock Domain adaptation via low rank and class discriminative representation for autism spectrum disorder identification: A multi-site fmri study.
\newblock \emph{IEEE Transactions on Neural Systems and Rehabilitation Engineering}, 31:\penalty0 806--817, 2023.
\newblock \doi{10.1109/TNSRE.2022.3233656}.

\bibitem[Maddison et~al.(2022)Maddison, Mnih, and Teh]{maddison2022concrete}
Chris~J Maddison, Andriy Mnih, and Yee~Whye Teh.
\newblock The concrete distribution: A continuous relaxation of discrete random variables.
\newblock In \emph{International Conference on Learning Representations}, 2022.

\bibitem[McGeown et~al.(2009)McGeown, Shanks, Forbes-McKay, and Venneri]{mcgeown2009patterns}
William~Jonathan McGeown, Michael~Fraser Shanks, Katrina~Elaine Forbes-McKay, and Annalena Venneri.
\newblock Patterns of brain activity during a semantic task differentiate normal aging from early alzheimer's disease.
\newblock \emph{Psychiatry Research: Neuroimaging}, 173\penalty0 (3):\penalty0 218--227, 2009.

\bibitem[Miao et~al.(2022)Miao, Liu, and Li]{miao2022interpretable}
Siqi Miao, Mia Liu, and Pan Li.
\newblock Interpretable and generalizable graph learning via stochastic attention mechanism.
\newblock In \emph{International Conference on Machine Learning}, pp.\  15524--15543. PMLR, 2022.

\bibitem[Paszke et~al.(2017)Paszke, Gross, Chintala, Chanan, Yang, DeVito, Lin, Desmaison, Antiga, and Lerer]{paszke2017automatic}
Adam Paszke, Sam Gross, Soumith Chintala, Gregory Chanan, Edward Yang, Zachary DeVito, Zeming Lin, Alban Desmaison, Luca Antiga, and Adam Lerer.
\newblock Automatic differentiation in pytorch.
\newblock 2017.

\bibitem[Pedregosa et~al.(2011)Pedregosa, Varoquaux, Gramfort, Michel, Thirion, Grisel, Blondel, Prettenhofer, Weiss, Dubourg, et~al.]{pedregosa2011scikit}
Fabian Pedregosa, Ga{\"e}l Varoquaux, Alexandre Gramfort, Vincent Michel, Bertrand Thirion, Olivier Grisel, Mathieu Blondel, Peter Prettenhofer, Ron Weiss, Vincent Dubourg, et~al.
\newblock Scikit-learn: Machine learning in python.
\newblock \emph{the Journal of machine Learning research}, 12:\penalty0 2825--2830, 2011.

\bibitem[Poldrack et~al.(2009)Poldrack, Halchenko, and Hanson]{poldrack2009decoding}
Russell~A Poldrack, Yaroslav~O Halchenko, and Stephen~Jos{\'e} Hanson.
\newblock Decoding the large-scale structure of brain function by classifying mental states across individuals.
\newblock \emph{Psychological science}, 20\penalty0 (11):\penalty0 1364--1372, 2009.

\bibitem[Sagawa et~al.(2019)Sagawa, Koh, Hashimoto, and Liang]{sagawa2019distributionally}
Shiori Sagawa, Pang~Wei Koh, Tatsunori~B Hashimoto, and Percy Liang.
\newblock Distributionally robust neural networks for group shifts: On the importance of regularization for worst-case generalization.
\newblock \emph{arXiv preprint arXiv:1911.08731}, 2019.

\bibitem[Schaefer et~al.(2018)Schaefer, Kong, Gordon, Laumann, Zuo, Holmes, Eickhoff, and Yeo]{schaefer2018local}
Alexander Schaefer, Ru~Kong, Evan~M Gordon, Timothy~O Laumann, Xi-Nian Zuo, Avram~J Holmes, Simon~B Eickhoff, and BT~Thomas Yeo.
\newblock Local-global parcellation of the human cerebral cortex from intrinsic functional connectivity mri.
\newblock \emph{Cerebral cortex}, 28\penalty0 (9):\penalty0 3095--3114, 2018.

\bibitem[Scherf et~al.(2015)Scherf, Elbich, Minshew, and Behrmann]{scherf2015individual}
K~Suzanne Scherf, Daniel Elbich, Nancy Minshew, and Marlene Behrmann.
\newblock Individual differences in symptom severity and behavior predict neural activation during face processing in adolescents with autism.
\newblock \emph{NeuroImage: Clinical}, 7:\penalty0 53--67, 2015.

\bibitem[Smith et~al.(2011)Smith, Miller, Salimi-Khorshidi, Webster, Beckmann, Nichols, Ramsey, and Woolrich]{smith2011network}
Stephen~M Smith, Karla~L Miller, Gholamreza Salimi-Khorshidi, Matthew Webster, Christian~F Beckmann, Thomas~E Nichols, Joseph~D Ramsey, and Mark~W Woolrich.
\newblock Network modelling methods for fmri.
\newblock \emph{Neuroimage}, 54\penalty0 (2):\penalty0 875--891, 2011.

\bibitem[Sun \& Saenko(2016)Sun and Saenko]{sun2016deep}
Baochen Sun and Kate Saenko.
\newblock Deep coral: Correlation alignment for deep domain adaptation.
\newblock In \emph{Computer Vision--ECCV 2016 Workshops: Amsterdam, The Netherlands, October 8-10 and 15-16, 2016, Proceedings, Part III 14}, pp.\  443--450. Springer, 2016.

\bibitem[Veli{\v{c}}kovi{\'c} et~al.(2017)Veli{\v{c}}kovi{\'c}, Cucurull, Casanova, Romero, Lio, and Bengio]{velivckovic2017graph}
Petar Veli{\v{c}}kovi{\'c}, Guillem Cucurull, Arantxa Casanova, Adriana Romero, Pietro Lio, and Yoshua Bengio.
\newblock Graph attention networks.
\newblock \emph{arXiv preprint arXiv:1710.10903}, 2017.

\bibitem[Venneri et~al.(2008)Venneri, McGeown, Hietanen, Guerrini, Ellis, and Shanks]{venneri2008anatomical}
Annalena Venneri, William~J McGeown, Heidi~M Hietanen, Chiara Guerrini, Andrew~W Ellis, and Michael~F Shanks.
\newblock The anatomical bases of semantic retrieval deficits in early alzheimer's disease.
\newblock \emph{Neuropsychologia}, 46\penalty0 (2):\penalty0 497--510, 2008.

\bibitem[Wang et~al.(2022)Wang, Yao, Ma, and Liu]{wang2022multi}
Nan Wang, Dongren Yao, Lizhuang Ma, and Mingxia Liu.
\newblock Multi-site clustering and nested feature extraction for identifying autism spectrum disorder with resting-state fmri.
\newblock \emph{Medical image analysis}, 75:\penalty0 102279, 2022.

\bibitem[Wang et~al.(2023)Wang, Chen, Dai, Xin, Gu, and Yu]{wang2023effective}
Xinlei Wang, Jinyi Chen, Bing~Tian Dai, Junchang Xin, Yu~Gu, and Ge~Yu.
\newblock Effective graph kernels for evolving functional brain networks.
\newblock In \emph{Proceedings of the Sixteenth ACM International Conference on Web Search and Data Mining}, pp.\  150--158, 2023.

\bibitem[Wang et~al.(2015)Wang, Wang, Zhang, Mchugh, Sun, Li, and Yang]{wang2015interhemispheric}
Zhiqun Wang, Jianli Wang, Han Zhang, Robert Mchugh, Xiaoyu Sun, Kuncheng Li, and Qing~X Yang.
\newblock Interhemispheric functional and structural disconnection in alzheimer’s disease: a combined resting-state fmri and dti study.
\newblock \emph{PLoS One}, 10\penalty0 (5):\penalty0 e0126310, 2015.

\bibitem[Worsley et~al.(2002)Worsley, Liao, Aston, Petre, Duncan, Morales, and Evans]{worsley2002general}
Keith~J Worsley, Chien~Heng Liao, John Aston, V~Petre, GH~Duncan, F~Morales, and Alan~C Evans.
\newblock A general statistical analysis for fmri data.
\newblock \emph{Neuroimage}, 15\penalty0 (1):\penalty0 1--15, 2002.

\bibitem[Wu et~al.(2020)Wu, Ren, Li, and Leskovec]{wu2020graph}
Tailin Wu, Hongyu Ren, Pan Li, and Jure Leskovec.
\newblock Graph information bottleneck.
\newblock \emph{Advances in Neural Information Processing Systems}, 33:\penalty0 20437--20448, 2020.

\bibitem[Wu et~al.(2022)Wu, Wang, Zhang, He, and seng Chua]{wu2022dir}
Ying-Xin Wu, Xiang Wang, An~Zhang, Xiangnan He, and Tat seng Chua.
\newblock Discovering invariant rationales for graph neural networks.
\newblock In \emph{ICLR}, 2022.

\bibitem[Xu et~al.(2023)Xu, Yang, Huang, Gururajapathy, Ke, Qiao, Wang, Kumar, McGeown, and Kwon]{xu2023data}
Jiaxing Xu, Yunhan Yang, David Tse~Jung Huang, Sophi~Shilpa Gururajapathy, Yiping Ke, Miao Qiao, Alan Wang, Haribalan Kumar, Josh McGeown, and Eryn Kwon.
\newblock Data-driven network neuroscience: On data collection and benchmark.
\newblock In \emph{Thirty-seventh Conference on Neural Information Processing Systems Datasets and Benchmarks Track}, 2023.

\bibitem[Xu et~al.(2024{\natexlab{a}})Xu, Bian, Li, Zhang, Ke, Qiao, Zhang, Sim, and Guly{\'a}s]{xu2024contrastive}
Jiaxing Xu, Qingtian Bian, Xinhang Li, Aihu Zhang, Yiping Ke, Miao Qiao, Wei Zhang, Wei Khang~Jeremy Sim, and Bal{\'a}zs Guly{\'a}s.
\newblock Contrastive graph pooling for explainable classification of brain networks.
\newblock \emph{IEEE Transactions on Medical Imaging}, 2024{\natexlab{a}}.

\bibitem[Xu et~al.(2024{\natexlab{b}})Xu, He, Lan, Bian, Li, Li, Ke, and Qiao]{xu2024contrasformer}
Jiaxing Xu, Kai He, Mengcheng Lan, Qingtian Bian, Wei Li, Tieying Li, Yiping Ke, and Miao Qiao.
\newblock Contrasformer: A brain network contrastive transformer for neurodegenerative condition identification.
\newblock In \emph{Proceedings of the 33rd ACM International Conference on Information and Knowledge Management}, pp.\  2671--2681, 2024{\natexlab{b}}.

\bibitem[Xu et~al.(2024{\natexlab{c}})Xu, Lan, Dong, He, Zhang, Bian, and Ke]{xu2024multi}
Jiaxing Xu, Mengcheng Lan, Xia Dong, Kai He, Wei Zhang, Qingtian Bian, and Yiping Ke.
\newblock Multi-atlas brain network classification through consistency distillation and complementary information fusion.
\newblock \emph{arXiv preprint arXiv:2410.08228}, 2024{\natexlab{c}}.

\bibitem[Xu et~al.(2018)Xu, Hu, Leskovec, and Jegelka]{xu2018powerful}
Keyulu Xu, Weihua Hu, Jure Leskovec, and Stefanie Jegelka.
\newblock How powerful are graph neural networks?
\newblock \emph{arXiv preprint arXiv:1810.00826}, 2018.

\bibitem[Xu et~al.(2021)Xu, Sanz, Garces, Maestu, Li, and Pantazis]{xu2021graph}
Mengjia Xu, David~Lopez Sanz, Pilar Garces, Fernando Maestu, Quanzheng Li, and Dimitrios Pantazis.
\newblock A graph gaussian embedding method for predicting alzheimer's disease progression with meg brain networks.
\newblock \emph{IEEE Transactions on Biomedical Engineering}, 68\penalty0 (5):\penalty0 1579--1588, 2021.

\bibitem[Yao et~al.(2024)Yao, Chen, Chen, Hu, Shen, and Zhang]{yao2024empowering}
Tianjun Yao, Yongqiang Chen, Zhenhao Chen, Kai Hu, Zhiqiang Shen, and Kun Zhang.
\newblock Empowering graph invariance learning with deep spurious infomax.
\newblock In \emph{Forty-first International Conference on Machine Learning}, 2024.

\bibitem[Yeo et~al.(2011)Yeo, Krienen, Sepulcre, Sabuncu, Lashkari, Hollinshead, Roffman, Smoller, Z{\"o}llei, Polimeni, et~al.]{yeo2011organization}
BT~Thomas Yeo, Fenna~M Krienen, Jorge Sepulcre, Mert~R Sabuncu, Danial Lashkari, Marisa Hollinshead, Joshua~L Roffman, Jordan~W Smoller, Lilla Z{\"o}llei, Jonathan~R Polimeni, et~al.
\newblock The organization of the human cerebral cortex estimated by intrinsic functional connectivity.
\newblock \emph{Journal of neurophysiology}, 2011.

\bibitem[Zhang et~al.(2022)Zhang, Song, Wang, Zhang, Wang, Wang, and Zhang]{zhang2022classification}
Hao Zhang, Ran Song, Liping Wang, Lin Zhang, Dawei Wang, Cong Wang, and Wei Zhang.
\newblock Classification of brain disorders in rs-fmri via local-to-global graph neural networks.
\newblock \emph{IEEE Transactions on Medical Imaging}, 2022.

\bibitem[Zhang et~al.(2018)Zhang, Cisse, Dauphin, and Lopez-Paz]{zhang2018mixup}
Hongyi Zhang, Moustapha Cisse, Yann~N. Dauphin, and David Lopez-Paz.
\newblock mixup: Beyond empirical risk minimization.
\newblock \emph{International Conference on Learning Representations}, 2018.
\newblock URL \url{https://openreview.net/forum?id=r1Ddp1-Rb}.

\end{thebibliography}
\bibliographystyle{iclr2025_conference}

\newpage

\appendix

\begin{center}
    \LARGE \bf {Appendix of BrainOOD}
\end{center}
\etocdepthtag.toc{mtappendix}
\etocsettagdepth{mtchapter}{none}
\etocsettagdepth{mtappendix}{subsection}
\tableofcontents

\clearpage

\section{Notation}
\label{app:notation}

Notation-wise, we use calligraphic letters to denote sets (e.g., $\mathcal{X}$), bold capital letters to denote matrices (e.g., $\boldsymbol{X}$), and strings with bold lowercase letters to represent vectors (e.g., $\boldsymbol{x}$).  Subscripts and superscripts are used to distinguish between different variables or parameters, and lowercase letters denote scalars. We use $\boldsymbol{S}[i,:]$ and $\boldsymbol{S}[:,j]$ to denote the $i$-th row and $j$-th column of a matrix $\boldsymbol{S}$, respectively. 
Table \ref{tab:notation} summarizes the notations used throughout the paper.

\begin{table}[h]
\centering
\caption{Notation table}
\begin{tabular}{cl}
\hline
Notation                         & Description                       \\ \hline
$\boldsymbol{S}$         &  A connectivity matrix       \\
$G$    &  A brain network \\
$G_C$    &  The causal subgraph for a brain network $G$ \\
$\boldsymbol{X}$         &  The feature matrix of a brain network        \\
$\boldsymbol{A}$         &  The adjacency matrix of a brain network        \\
$\mathcal{D}$               & Input dataset         \\
$\mathcal{Y}$              & Input label set             \\
$y_G$                            & Label of brain network $G$  \\
$n$    & Number of nodes/ROIs \\     
$\boldsymbol{H}_v$                & Node representation of $v$         \\
$d$        & Dimensionality of node representations \\
$\mathcal{G}$ & The graph space \\
$\mathcal{G}_C$  &  The space of subgraphs with respect to the graphs from $\mathcal{G}$ \\
$\boldsymbol{W}_{mask}$                 & Parameter matrices   \\
$\boldsymbol{M}$    & The learnable mask             \\
$\boldsymbol{X}^{\prime}$  &  The masked node feature matrix  \\
$\hat{\boldsymbol{H}}$  &  The recovered node representations  \\
$\hat{\boldsymbol{X}}$  &  The recovered node features with mask \\
$i, j, v, u$            & Index for matrix dimensions            \\ 
$\boldsymbol{A}^{\prime}$  &  The sampled adjacency matrix  \\
$\alpha_{v, u}$  & The score of edge $(v, u)$  \\ 
$\gamma_{v, u}$ & The sampling probability for edge $(v, u)$ \\
$\sigma^\prime$  &  The standard deviation matrix of all the $\boldsymbol{A}^\prime$ in a batch  \\
$g_\phi$                            & The subgraph extractor with parameter $\phi$\\ &to generate a subgraph $G_C$ to interpret brain network $G$  \\
$I(\cdot;\cdot)$                            & Mutual information  \\
$H(\cdot)$                            & Entropy  \\
$\hat{y}_G$                            & The final prediction of brain network $G$  \\\hline
\end{tabular}
\label{tab:notation}
\end{table}

\section{Theoretical Discussion and Proofs}
\subsection{Proof for Theorem~\ref{thm:gib_loss}}
\label{proof:gib_loss}
\begin{theorem}[Restatement of Theorem~\ref{thm:gib_loss}]
\label{thm:gib_loss_appdx}
    For a subgraph extractor $g_\phi$ that encodes the input graph $G$ into representation $\mH$ to extract the desired subgraph $G_C^*$, if $g_\phi$ is limited in representation power, i.e., $I(G;\mH)< H(G_C^*)$,  where $H(\cdot)$ is the entropy of the underlying causal subgraph $G_C^*$, then solving for GIB objective:
    \begin{equation}\label{eq:gib_appdx}
    \text{$\max$}_{G_C}I(G_C; y_G)-\beta I(G_C;G),\ G_C\sim g_\phi(G),
\end{equation} can not elicit $G_C^*$.
\end{theorem}
\begin{proof}
    Given the GIB objective, 
    following previous works~\citep{miao2022interpretable,cheninterpretable}, we have:
    \begin{equation}\label{eq:gib_sol}
    \begin{aligned}
        I(G_C;y_G)-\beta I(G_C;G)&=I(y_G;G,G_C)-I(G;y_G|G_C)-\beta I(G_C;G)\\
        &=I(y_G;G,G_C)-(1-\beta)I(G;y_G|G_C)-\beta I(G;G_C,y_G)\\
        &=(1-\beta)I(y_G;G)-(1-\beta)I(G;y_G|G_C)-\beta I(G;G_C|y_G).
    \end{aligned}
    \end{equation}
    Since $I(y_G;G)$ is fixed given the data generation process, maximizing Eq.~(\ref{eq:gib_sol}) is equivalent to minimize $(1-\beta)I(G;y_G|G_C)-\beta I(G;G_C|y_G)$.
    The minimizer is taken and only taken when $G_C=G_C^*$.
    
    However, given the subgraph extractor $g_\phi$ that encodes the input graph $G$ into representation $\mH$ to extract the desired subgraph $G_C^*$, we have a Markov chain $G_C^*\rightarrow G\rightarrow \mH\rightarrow G_C$, from which we know that
    \begin{equation}
        I(G_C;G_C^*)\leq I(G;\mH).
    \end{equation}
    If $g_\phi$ is limited in representation is lower, i.e., $I(G;\mH)< H(G_C^*)$, then it suffices to know that $I(G_C;G_C^*)<H(G_C^*)$, and $G_C\neq G_C^*$.
\end{proof}

\section{More Details about Datasets}

\subsection{Detailed Dataset Description}
\label{app:dataset}

The class-wise sample sizes are summarized in Table \ref{tab:dataset_class_distribution}.

\begin{table}[h]
    \caption{The Class Distribution of the Brain Network Datasets we used}
    \centering
    \begin{tabular}{ccccc}
    \hline
       Dataset         & Gender (F/M) &  Age (mean ± std)      & Class   & \# Subjects \\
    \hline
        \multirow{2}{*}{ABIDE} & \multirow{2}{*}{152/873} & \multirow{2}{*}{16.5 ± 7.4}    & Control    & 537 \\
                             & &     & ASD        & 488 \\
    \hline
        \multirow{6}{*}{ADNI}   & \multirow{6}{*}{728/599} & \multirow{6}{*}{74.6 ± 7.9}    & CN    & 819 \\
                               & &   & SMC        & 73  \\
                              & &    & LMCI       & 102 \\
                              & &    & MCI       & 179 \\
                               & &   & EMCI       & 89  \\
                               & &   & AD         & 65  \\
    \hline
    \end{tabular}
    \label{tab:dataset_class_distribution}
\end{table}

\textbf{ABIDE} The ABIDE initiative supports the research on ASD by aggregating functional brain imaging data from laboratories worldwide. ASD is characterized by stereotyped behaviors, including irritability, hyperactivity, depression, and anxiety. Subjects in the dataset are classified into two groups: TC and individuals diagnosed with ASD. 

\textbf{ADNI} The ADNI raw images used in this paper were obtained from the ADNI database (\url{adni.loni.usc.edu}). The ADNI was launched in 2003 as a public-private partnership, led by Principal Investigator Michael W. Weiner, MD. The primary goal of ADNI has been to test whether serial magnetic resonance imaging (MRI), PET, other biological markers, and clinical and neuropsychological assessment can be combined to measure the progression of mild cognitive impairment (MCI) and early AD. For up-to-date information, see \url{www.adni-info.org}. We include subjects from 6 different stages of AD, from cognitive normal (CN), significant memory concern (SMC), mild cognitive impairment (MCI), early MCI (EMCI), late MCI (LMCI) to AD.

\subsection{Detailed Data Splits under OOD Setting}
\label{app:split}

Table \ref{tab:site} provides detailed information on the specific sites and the number of subjects used as OOD set in each fold. With such data split, the proportion of OOD subjects in the test set of each fold is in the range of [30\%, 55\%]. Subjects from the other sites are evenly assigned to each fold.

For the ABIDE dataset, given that the average number of subjects per site is approximately 60, we selected the smallest 10 sites as OOD sets across the 10 folds. This ensures that the test sets in all folds contain a mixture of both ID and OOD subjects, allowing for a robust evaluation of the model's generalization capabilities.

In contrast, for the ADNI dataset, where the number of sites is larger and the average number of subjects per site is only around 22, we selected the largest 10 sites as OOD sets across the 10 folds. This choice ensures that there are enough OOD subjects in the test set of each fold to reliably assess the model's performance under OOD conditions.

\begin{table}[h]
\centering
\caption{The Site Chosen as OOD set in Each Fold of ABIDE and ADNI Datasets.}
\label{tab:site}
\begin{tabular}{c|cc|cc}
\hline
\multirow{2}{*}{Fold} & \multicolumn{2}{c|}{ABIDE} & \multicolumn{2}{c}{ADNI} \\
                      & Site Name    & Subject\#   & SITEID    & Subject\#    \\ \hline
1                     & SBL          & 30          & 58        & 73           \\
2                     & OLIN         & 36          & 59        & 62           \\
3                     & SDSU         & 36          & 20        & 57           \\
4                     & CALTECH      & 38          & 27        & 50           \\
5                     & STANFORD     & 40          & 52        & 50           \\
6                     & TRINITY      & 49          & 47        & 46           \\
7                     & KKI          & 55          & 2         & 46           \\
8                     & YALE         & 56          & 25        & 45           \\
9                     & MAX\_MUN     & 57          & 5         & 43           \\
10                    & PITT         & 57          & 1         & 39           \\ \hline
\end{tabular}
\end{table}

\section{More Details about the Experiments}

\subsection{Baseline Descriptions}
\label{app:baseline}

\begin{itemize}
    \item General OOD Methods. \\
    \textbf{ERM}~\citep{goyal2017accurate}: Empirical Risk Minimization, which trains on the full dataset without specific domain adaptation. \\
    \textbf{Deep Coral}~\citep{sun2016deep}: Minimizes the domain shift by aligning covariance matrices across domains.\\
    \textbf{IRM}~\citep{arjovsky2019invariant}: Seeks to find invariant features across different environments by penalizing variations. \\
    \textbf{GroupDRO}~\citep{sagawa2019distributionally}: Tackles minority distributions by optimizing the worst-case group performance.  \\
    \textbf{VREx}~\citep{krueger2021out}: Reduces the risk variance across training environments to improve robustness.
    \item Graph OOD Methods. \\
    \textbf{Mixup}~\citep{zhang2018mixup}: Trains the model on convex combinations of pairs of examples to enhance robustness. \\
    \textbf{DIR}~\citep{wu2022dir}: Selects causal subgraphs and conducts interventional augmentation to enhance OOD generalization. \\
    \textbf{GSAT}~\citep{miao2022interpretable}: Incorporates stochasticity in attention weights to filter task-irrelevant subgraphs while enhancing interpretability. \\
    \textbf{GMT}~\citep{cheninterpretable}: Extracts interpretable subgraphs via approximation methods to achieve OOD generalization.
    \item General-Purpose GNNs. \\
    \textbf{GCN}~\citep{kipf2016semi}: A Graph Convolutional Network baseline with mean pooling. \\
    \textbf{GIN}~\citep{xu2018powerful}: A Graph Isomorphism Network with sum pooling, which adjusts node importance using learnable parameters. \\
    \textbf{GAT}~\citep{velivckovic2017graph}: A Graph Attention Network, which applies attention mechanisms to learn node-to-neighbor importance weights.
    \item Neural Networks Tailored for Brain Networks. \\
    \textbf{BrainNetCNN}~\citep{kawahara2017brainnetcnn}: A Convolutional Neural Network developed for connectome data.\\
    \textbf{BrainGNN}~\citep{li2021braingnn}: A GNN-based method that incorporates ROI-aware convolution layers for integrating fMRI data. \\
    \textbf{ContrastPool}~\citep{xu2024contrastive}: A pooling method that clusters nodes and uses dual-attention mechanisms for domain-specific information. \\
    \textbf{Contrasformer}~\citep{xu2024contrasformer}: A transformer-based approach with contrastive constraints applied at both ROI and population levels.
\end{itemize}

\subsection{Implementation Details}
\label{app:imple_detail}

For all OOD methods, we use the same GNN architecture as graph encoders, following GSAT~\citep{miao2022interpretable}.We use 2-layer GIN~\citep{xu2018powerful} with Batch Normalization \citep{10.5555/3045118.3045167} as the backbone. The hidden dimension is set to 100 and the dropout ratio is set to 0.5. The pooling function is sum pooling. The settings of our experiments about OOD methods follow those in GOOD~\citep{gui2022good}. The whole network is trained in an end-to-end manner using the Adam optimizer \citep{kingma2014adam} with a learning rate of 1e-3 and a batch size of 64 for all OOD models at all datasets. All OOD models are trained for 100 epochs. The final model is selected according to the best validation classification performance on ID and OOD sets, respectively. We report the mean and standard deviation of 10 folds to evaluate how these models can generalize to the unseen OOD sites. All the codes were implemented using PyTorch~\citep{paszke2017automatic} and PyTorch Geometric~\citep{Fey/Lenssen/2019} packages. The optimized hyperparameters for BrainOOD are reported in Table~\ref{tab:optimized_params}.

\begin{table}[h]
\caption{The optimized hyperparameters for BrainOOD.}
\centering
\begin{tabular}{ccc}
\hline
             & ABIDE & ADNI  \\ \hline
feature dropout     & 0.2    & 0.2    \\
$\lambda_1$    & 0.01   & 0.01   \\
$\lambda_2$    & 0.1  & 10    \\
$\lambda_3$    & 0.5  & 0.1  \\
$k$     & 5  & 3 \\ \hline
\end{tabular}
\label{tab:optimized_params}
\end{table}

The experiments of general-purposed GNNs and models tailored for brain networks based on the framework used in ContrastPool \citep{xu2024contrastive}. The learning rate and batch size are using author-recommended values for fair comparison. The maximum number of training epochs is set to 1000. We use the early stopping criterion, i.e., we stop the training once there is no further improvement on the validation loss during 25 epochs. The whole network is trained in an end-to-end manner using the Adam optimizer \citep{kingma2014adam} with.

All experiments were conducted on a Linux server with an Intel(R) Core(TM) i9-10940X CPU (3.30GHz), a GeForce GTX 3090 GPU, and a 125GB RAM.

\section{More Experimental Results}

\subsection{In-depth Analysis for the Performance on Different Sites}
\label{app:ood_acc}

\begin{figure}[h]
\centering
\includegraphics[width=.9\textwidth]{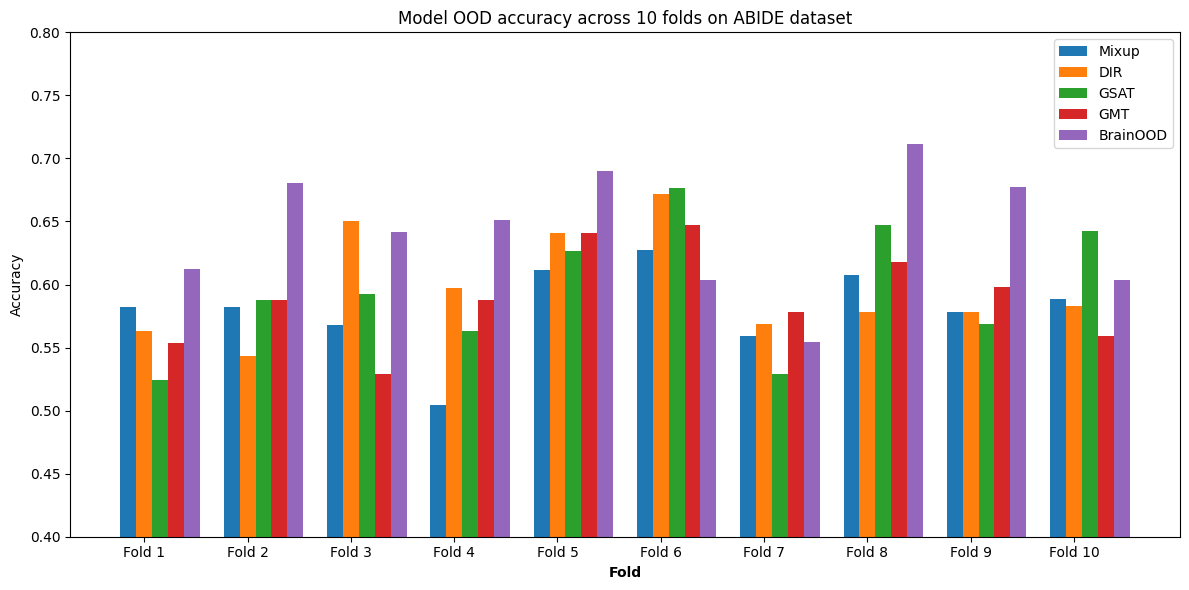}
\includegraphics[width=.9\textwidth]{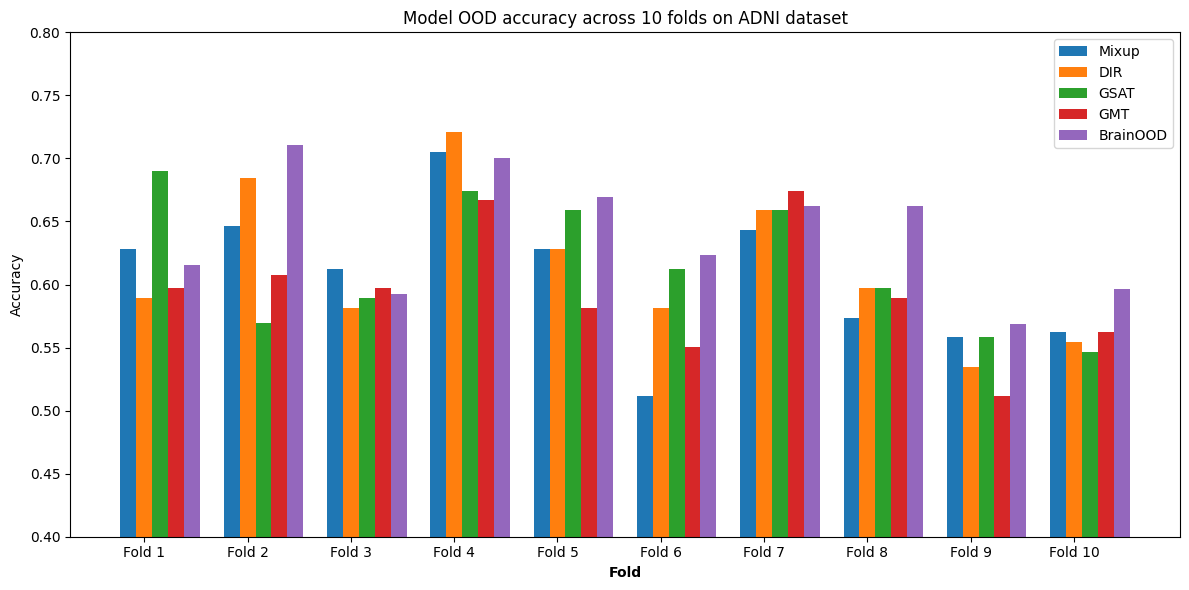}
\caption{Comparison with graph OOD methods in terms of test OOD accuracy across 10 folds on ABIDE and ADNI datasets.}
\label{fig:ood_acc}
\end{figure}

We also conducted a detailed evaluation of the OOD set in each fold, which reveals how well the models generalize to unseen sites. Figure \ref{fig:ood_acc} presents a comparison of BrainOOD against four other graph OOD methods. The trends across different folds on the two datasets are consistent, and we observe large variances for accuracy across different folds, especially on ADNI dataset. This indicates that some sites are significantly different from others, making it difficult for models to generalize effectively to these sites.

On the ABIDE dataset, BrainOOD achieves the best results on 6 out of 10 folds and secures the second-best performance on 2 other folds. BrainOOD surpasses the runner-up model up to 10\% (on fold 2). Similarly, on the ADNI dataset, BrainOOD also ranks first on 6 out of 10 folds and second-best on 2 additional folds. BrainOOD surpasses the runner-up model up to 6\% (on fold 8). Notably, BrainOOD never ranks as the worst-performing model across all folds of both datasets. The worst performance for BrainOOD is still the best compared to the worst one of other models on all folds of these datasets.

These results demonstrate that BrainOOD not only has strong generalization capabilities but also exhibits robustness in its performance across multiple unseen sites, making it a reliable choice for OOD scenarios in brain network analysis.

\subsection{Experimental Results with Other Backbone}
\label{app:backbone}

To verify the adaptability of the BrainOOD framework to different GNN backbones, we conducted experiments by integrating various graph OOD methods with GCN backbones. The results, presented in Table \ref{tab:gcn_ood}, demonstrate that existing OOD methods fail to improve performance when combined with the GCN backbone, emphasizing the necessity of designing OOD algorithms specifically tailored for brain networks.

In contrast, integrating BrainOOD with the GCN backbone results in a notable improvement, achieving a 6.3\% increase in overall accuracy. This significant gain highlights the effectiveness of BrainOOD in enhancing the generalization capabilities of GNN models for brain network analysis, even when applied to general-purpose backbones like GCN.

\begin{table}[h]
\setlength\tabcolsep{5pt}
\centering
\small
\caption{Results of graph OOD methods with GCN backbone. The best result is highlighted in \textbf{bold}.}
\label{tab:gcn_ood}
\begin{tabular}{l|ccc|ccc}
\hline
\multirow{2}{*}{Model} & \multicolumn{3}{c|}{ABIDE}                 & \multicolumn{3}{c}{ADNI (6-class)}                     \\
                                 & ID acc       & OOD acc      & Overall acc  & ID acc        & OOD acc       & Overall acc  \\ \hline
GCN  & - & - & 61.85 ± 4.39 & - & -  & 60.92 ± 4.13 \\\hline
Mixup                            & 60.78 ± 5.01 & 58.06 ± 6.06 & 59.52 ± 3.93 & 59.34 ± 7.52  & 60.01 ± 13.65 & 59.69 ± 5.37 \\
DIR                              & 60.66 ± 6.53 & 57.81 ± 5.56 & 59.76 ± 2.69 & 60.71 ± 10.04 & 60.20 ± 14.18 & 60.23 ± 5.05 \\
GSAT                             & 62.73 ± 4.47 & 59.12 ± 6.17 & 61.27 ± 2.03 & 58.67 ± 10.02 & 57.99 ± 15.37 & 57.89 ± 7.19 \\
GMT                              & 63.38 ± 5.23 & 58.14 ± 7.41 & 61.56 ± 4.05 & 60.34 ± 11.00 & 56.31 ± 11.28 & 58.68 ± 6.92 \\
BrainOOD                         & \textbf{64.91} ± 4.23 & \textbf{62.85} ± 6.88 & \textbf{63.34} ± 2.77 & \textbf{66.54} ± 11.51 & \textbf{62.05} ± 14.50 & \textbf{64.10} ± 5.16 \\ \hline
\end{tabular}
\end{table}

\subsection{Model Interpretation with ADNI}
\label{app:case_adni}

\begin{figure}[h]
\centering
\includegraphics[width=0.45\textwidth]{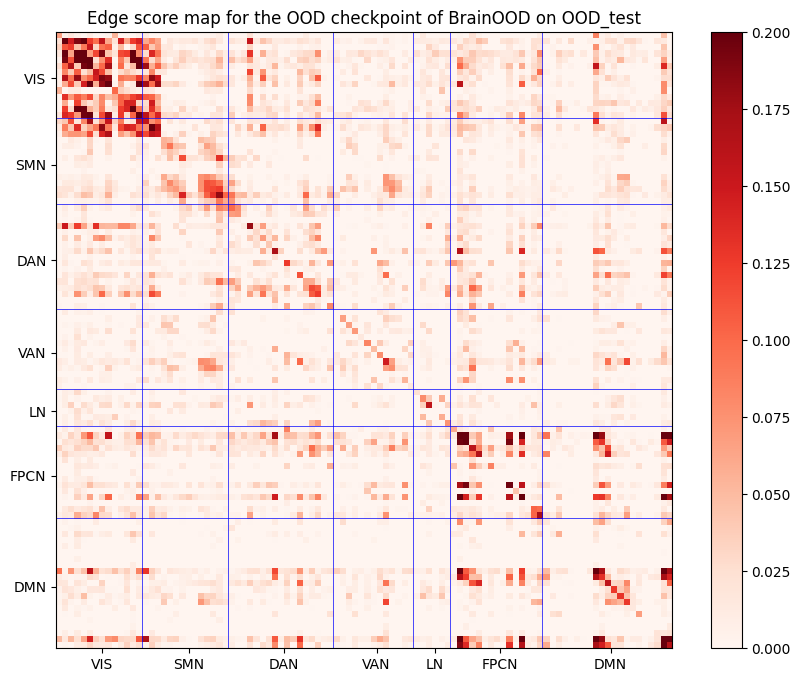}
\includegraphics[width=0.45\textwidth]{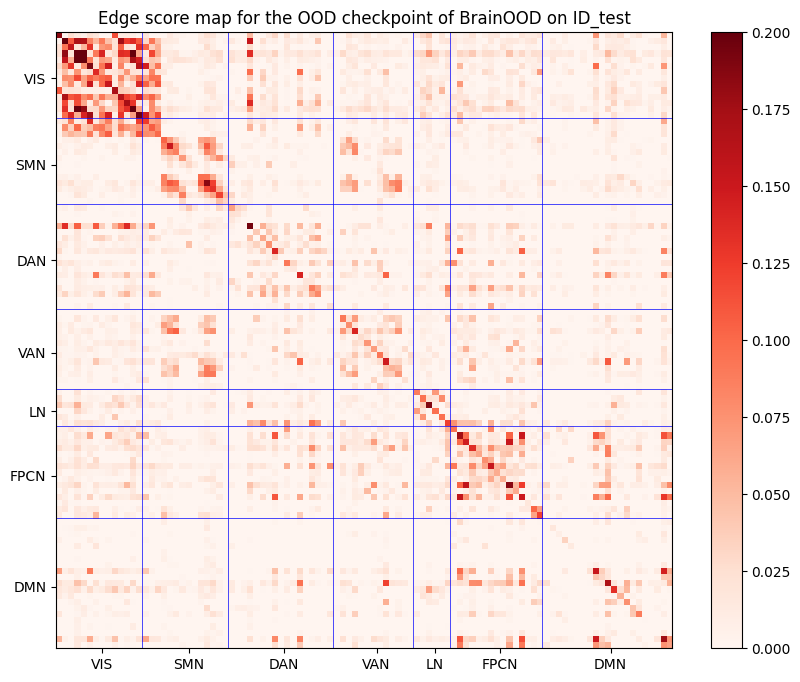}
\includegraphics[width=0.45\textwidth]{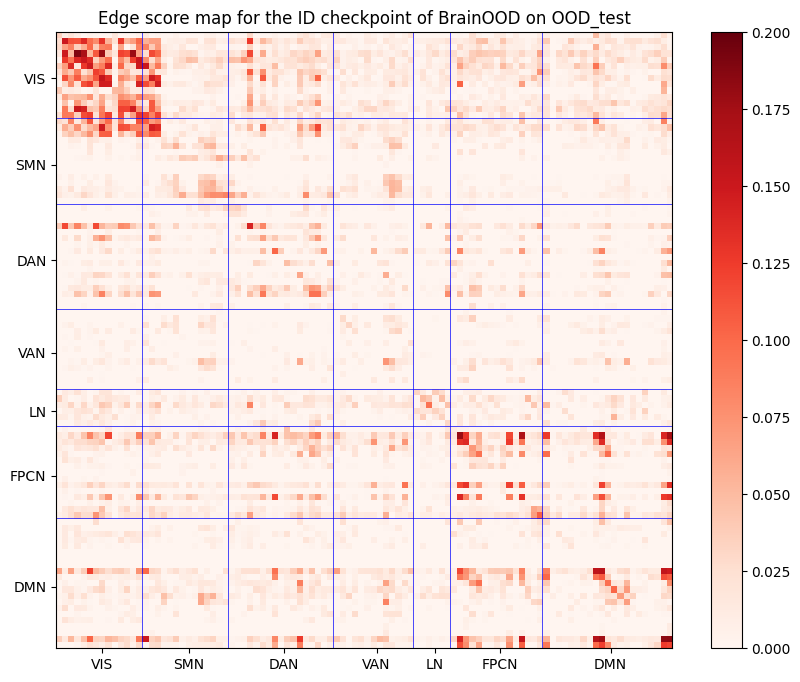}
\includegraphics[width=0.45\textwidth]{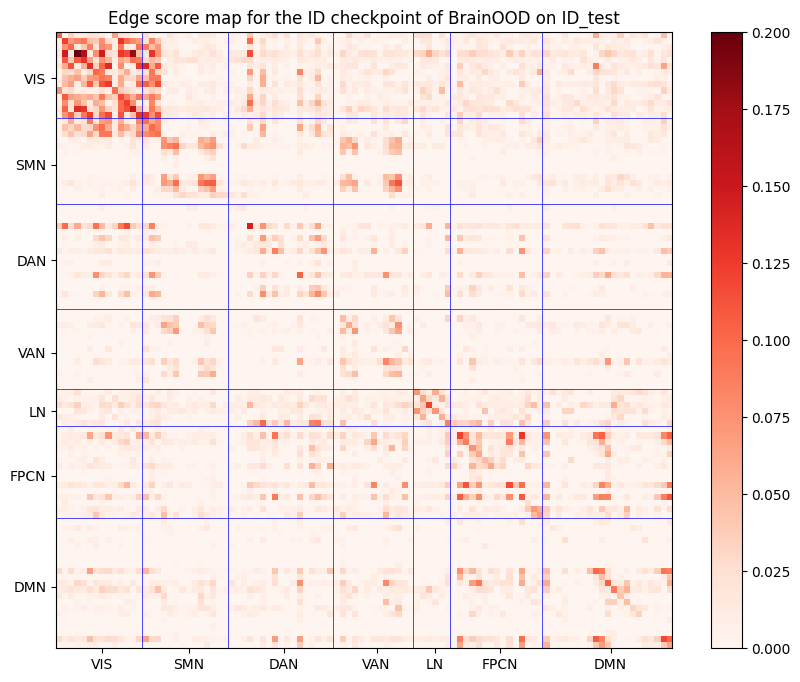}
\caption{Edge score map visualization for ADNI dataset. VIS = visual network; SMN = somatomotor network; DAN = dorsal attention network; VAN = ventral attention network; LN = limbic network; FPCN = frontoparietal control network; DMN = default mode network.}
\label{fig:vis_adni_brainood}
\end{figure}

In the ADNI dataset, we observed similar consistency between score maps for both ID and OOD test sets when evaluated using the same checkpoint, as illustrated in Figure \ref{fig:vis_abide_brainood}. This consistency once again highlights BrainOOD's ability to capture invariant patterns from OOD subjects.  When comparing different checkpoints on the same test sets, both ID and OOD checkpoints identify common connections within VIS and frontoparietal control network (FPCN), both of which are recognized as important connectivity regions in AD research~\citep{jiang2020stronger,boyle2024left}. Additionally, some connections, such as those within SMN, are uniquely highlighted in the OOD checkpoint, emphasizing the variations that may arise between the different test environments.

\begin{wrapfigure}{r}{0.4\textwidth}
\begin{center}
\includegraphics[width=\linewidth]{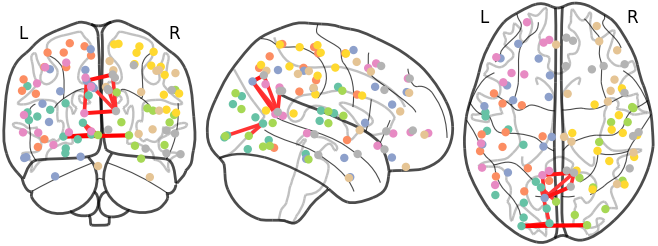}
\end{center}
\caption{The visualization of the top 10 connections with the highest score on ADNI OOD set.}
\label{fig:brain_adni}
\end{wrapfigure}

For the most significant connections in the causal subgraph of ADNI, we selected the top 10 connections with the highest scores, as shown in Figure \ref{fig:brain_adni}. These highlighted connections across the left and right hemispheres, particularly between the lateral prefrontal cortex and medial posterior prefrontal cortex, suggest potential AD-specific neural mechanisms. Previous studies have identified these regions as critical in AD progression~\citep{venneri2008anatomical,mcgeown2009patterns}. Notably, research also indicates that interhemispheric connectivity, particularly involving the corpus callosum, plays a crucial role in AD~\citep{wang2015interhemispheric}, further validating our model's interpretability in identifying AD-relevant neural patterns.

\subsection{Hyperparameter Analysis}
\label{app:hyper}

In this section, we study the sensitivity of three trade-off hyperparameters in Eq. (\ref{eq:total_loss}) and the sampling number $k$. All experiments are conducted on the ABIDE dataset. We tune the value of $\lambda_1$ from \{0.001, 0.01, 0.1\}, $\lambda_2$ from \{0.01, 0.1, 1.0\}, $\lambda_3$ from \{0.1, 0.5, 1.0\}, and $k$ from \{1, 3, 5, 10, 20\}. The results presented in Table \ref{tab:tune_lambda} show that our model performs the best when $\lambda_1=0.01$, $\lambda_2=0.1$, $\lambda_3=0.5$, and $k=5$. We can exhibit that the influence of $\lambda_1$ and $\lambda_2$ is larger than $\lambda_3$, which implies the importance of introducing feature selection with a suitable trade-off.

\begin{table}[h]
\caption{The hyperparameter sensitivity analysis for BrainOOD on ABIDE dataset.}
\centering
\label{tab:tune_lambda}
\begin{tabular}{ccccc}
\hline
$k$ & $\lambda_1$    & $\lambda_2$  & $\lambda_3$  & overall acc          \\ \hline
1 & 0.01 & 0.1  & 0.5 & 61.95 ± 4.54 \\ 
3 & 0.01 & 0.1  & 0.5 & 61.37 ± 3.38 \\ 
5 & 0.001 & 0.1  & 0.1 & 61.31 ± 5.26 \\ 
5 & 0.001 & 0.1  & 0.5 & 62.19 ± 3.45 \\
5 & 0.001 & 0.1  & 1.0 & 61.98 ± 5.56 \\
5 & 0.01  & 0.01 & 0.1 & 61.71 ± 3.49 \\
5 & 0.01  & 0.01 & 0.5 & 62.52 ± 4.15 \\
5 & 0.01  & 0.01 & 1.0 & 61.71 ± 3.49 \\
5 & 0.01  & 0.1  & 0.1 & 62.98 ± 3.57 \\
5 & 0.01  & 0.1  & 0.5 & \textbf{63.95} ± 4.65 \\ 
5 & 0.01  & 0.1  & 1.0 & 62.72 ± 4.00 \\
5 & 0.01  & 1.0  & 0.1 & 62.05 ± 5.14 \\
5 & 0.01  & 1.0  & 0.5 & 61.15 ± 2.84 \\
5 & 0.01  & 1.0  & 1.0 & 60.59 ± 5.24 \\
5 & 0.1   & 0.1  & 0.1 & 61.46 ± 4.41 \\
5 & 0.1   & 0.1  & 0.5 & 61.66 ± 3.65 \\
5 & 0.1   & 0.1  & 1.0 & 62.00 ± 4.50 \\ 
10 & 0.01 & 0.1  & 0.5 & 62.90 ± 4.67 \\ 
20 & 0.01 & 0.1 & 0.5 & 61.59 ± 3.57 \\ \hline
\end{tabular}
\end{table}

\section{More Related Works about Brain Network Analysis with GNNs}
\label{app:related}

In recent years, several GNN-based methods have been proposed for brain network analysis. \citet{ktena2017distance} leverages graph convolutional networks (GCNs) for learning similarities between each pair of graphs (subjects). BrainNetCNN \citep{kawahara2017brainnetcnn} proposes edge-to-edge, edge-to-node and node-to-graph convolutional filters to leverage the topological information of brain networks in the neural network. PRGNN \citep{li2020pooling} proposes a graph pooling method with group-level regularization to guarantee group-level consistency. BrainGNN \citep{li2021braingnn} proposes an ROI-selection pooling to highlight salient ROIs for each individual.  MG2G \citep{xu2021graph} is a two-stage approach. The first stage learns node representations through a self-supervised link prediction task. The second stage employs the learned representations to train a classifier for predicting Alzheimer's disease progression. LG-GNN \citep{zhang2022classification} incorporates local ROI-GNN and global subject-GNN guided by non-imaging data, such as gender, age, and acquisition site. Some more recent works \citep{xu2024contrastive,xu2024contrasformer} introduce a contrast graph to highlight the difference between groups and thus improve the model's generalization ability. Despite these advancements, addressing the OOD challenge in brain network analysis remains largely unexplored. 
Furthermore, while data harmonization methods \citep{guan2021multi,wang2022multi} and domain adaptation methods \citep{10198494,10012442} have been widely applied in study generalizing brain network models to other sites. However, these methods typically rely on learning a mapping from a source to a target domain, assuming the availability of the target domain distribution during training. In contrast, our study addresses the OOD generalization setting, where target domain data is entirely unseen during training. This stricter constraint represents a more challenging and realistic scenario, particularly in clinical applications where models must generalize to previously unseen sites without retraining. As a result, domain adaptation methods may be less effective in this context. Our work, therefore, pioneers the evaluation of brain network classification under an OOD generalization framework, emphasizing the need for new OOD algorithms specifically designed for brain networks.

\end{document}